\documentclass[11pt]{article}
\usepackage[a4paper,margin=1.2in]{geometry}

\usepackage[utf8]{inputenc}
\usepackage[T1]{fontenc}    %
\usepackage{hyperref}       %
\usepackage{url}            %
\usepackage{booktabs}       %
\usepackage{amsfonts}       %
\usepackage{amsmath,amssymb}
\usepackage{pifont}%
\newcommand{\cmark}{\ding{51}}%
\newcommand{\xmark}{\ding{55}}%
\usepackage{graphicx}
\usepackage{nicefrac}       %
\usepackage{microtype}      %
\usepackage[acronym, nopostdot, nogroupskip, nohypertypes={acronym,notation}]{glossaries}
\setacronymstyle{long-short}
\usepackage{lipsum}
\usepackage{pythonhighlight}
\usepackage{soul} %
\usepackage{rotating} %
\usepackage{algorithm}
\usepackage{subcaption}
\newtheorem{theorem}{Theorem}
\usepackage{algorithmic}
 \usepackage{setspace}
\let\Algorithm\algorithm
\renewcommand\algorithm[1][]{\Algorithm[#1]\setstretch{1.2}}

\usepackage{tikz}
\usetikzlibrary{arrows.meta,positioning}
\usepackage{xspace}
\usepackage{placeins}
\usepackage{natbib}

\newacronym{rkhs}{RKHS}{Reproducing Kernel Hilbert Space}
\newacronym{ipm}{IPM}{integral probability metric}
\newacronym{MCMC}{MCMC}{Markov chain Monte-Carlo}
\newacronym{HMC}{HMC}{Hamiltonian Monte-Carlo}
\newacronym{svi}{SVI}{stochastic variational inference}
\newacronym{svgd}{SVGD}{Stein Variational Gradient Descent}
\newacronym{VI}{VI}{variational inference}
\newacronym{kl}{KL}{Kullback-Leibler}
\newacronym{iid}{IID}{independent and identically distributed}
\newacronym{elbo}{ELBO}{evidence lower bound}
\newacronym{RBF}{RBF}{Radial Basis Function}
\newacronym{DPP}{DPP}{determinantal point process}
\newacronym{ARD}{ARD}{Automatic Relevance Determination }
\newacronym{KSD}{KSD}{Kernel Stein Discrepancy}
\newacronym{MMD}{MMD}{Maximum mean discrepancy}
\newacronym{MAP}{MAP}{maximum a posteriori}
\newacronym{NOx}{NOx}{nitrogen oxide}
\newacronym{NO2}{NO2}{Nitrogen Dioxide}
\newacronym{RMSE}{RMSE}{root-mean-square error}
\newacronym{UK}{UK}{United Kingdom}
\newacronym{tvd}{TVD}{total variation distance}
\newglossaryentry{GP}
{
  name={GP},
  description={Gaussian process},
  first={\glsentrydesc{GP} (\glsentrytext{GP})},
  plural={GPs},
  descriptionplural={Gaussian processes},
  firstplural={\glsentrydescplural{GP} (\glsentryplural{GP})}
} 
\newglossaryentry{DGP}
{
  name={DGP},
  description={deep Gaussian process},
  first={\glsentrydesc{DGP} (\glsentrytext{DGP})},
  plural={DGPs},
  descriptionplural={deep Gaussian processes},
  firstplural={\glsentrydescplural{DGP} (\glsentryplural{DGP})}
}

\newglossaryentry{RFF}
{
  name={RFF},
  description={Random Fourier Feature},
  first={\glsentrydesc{RFF} (\glsentrytext{RFF})},
  plural={RFF},
  descriptionplural={Random Fourier Features},
  firstplural={\glsentrydescplural{RFF} (\glsentryplural{RFF})}
} 

\usepackage{xcolor}

\definecolor{dkgreen}{rgb}{0,0.6,0}
\definecolor{gray}{rgb}{0.5,0.5,0.5}
\definecolor{mauve}{rgb}{0.58,0,0.82}

\definecolor{midnightblue}{rgb}{0.15, 0.15, 0.5}

\newcommand{\tabref}[1]{Table \ref{#1}}
\newcommand{\figref}[1]{Figure \ref{#1}}
\newcommand{\secref}[1]{Section \ref{#1}}
\newcommand{\algref}[1]{Algorithm \ref{#1}}
\newcommand{\appref}[1]{Appendix \ref{#1}}

\newcommand{\trace}{\text{trace}}

\newcommand{\norm}[1]{\left\lVert#1\right\rVert}

\newcommand{\x}{\mathbf{x}}

\newcommand{\by}{\mathbf{y}}

\DeclareMathOperator*{\argmin}{arg\,min}

\newcommand{\Kxx}{K_{\mathbf{xx}}}

\newcommand{\Kzz}{K_{\mathbf{Z Z}}}

\newcommand{\KL}{\text{KL}}

\newcommand{\varParam}{\xi}

\newcommand{\notwo}{NO\textsubscript{2}\xspace}

\newcommand{\bu}{\mathbf u}

\newcommand{\f}{\mathbf{f}}
\newcommand{\given}{\,|\,}
\newcommand{\btheta}{\boldsymbol\theta}

\newcommand{\particle}{\lambda}
\newcommand{\bparticle}{\boldsymbol\lambda}
\newcommand{\Particle}{\Lambda}

\newcommand{\clf}[1]{\textcolor{blue}{#1}}

\title{\textbf{Stein Variational Gaussian Processes}}
\author{Thomas Pinder$^{1}$, Christopher Nemeth$^1$, David Leslie$^1$ \\ \small
$^1$Department of Mathematics and Statistics, Lancaster University, UK 
}
\date{\today}

\begin{document}

\maketitle

\begin{abstract}
We show how to use Stein variational gradient descent (SVGD) to carry out inference in Gaussian process (GP) models with non-Gaussian likelihoods and large data volumes. Markov chain Monte Carlo (MCMC) is extremely computationally intensive for these situations, but the parametric assumptions required for efficient variational inference (VI) result in incorrect inference when they encounter the multi-modal posterior distributions that are common for such models. SVGD provides a non-parametric alternative to variational inference which is substantially faster than MCMC. %
We prove that for GP models with Lipschitz gradients the SVGD algorithm monotonically decreases the Kullback-Leibler divergence from the sampling distribution to the true posterior. Our method is demonstrated on benchmark problems in both regression and classification, a multimodal posterior, and an air quality example with 550,134 spatiotemporal observations, showing substantial performance improvements over MCMC and VI.
\end{abstract}

\section{Introduction}
\glspl{GP} are highly expressive, non-parametric distributions over continuous functions and are frequently employed in both regression and classification tasks \citep{rasmussen_gaussian_2006}. In recent years, \glspl{GP} have received significant attention in the machine learning community due to their successes in domains such as reinforcement learning \citep{deisenroth_gaussian_2015}, variance reduction \citep{oates_control_2017}, and optimisation \citep{mockus_bayesian_2012}. This recent blossoming has been facilitated by advances in inference methods, and especially by \gls{VI} which provides a tractable approach to fitting GP models to large and/or non-Gaussian data sets \citep[e.g.][]{hensman_gaussian_2013, cheng_variational_2017}.

While computationally efficient, \gls{VI} typically relies on the practitioner placing a parametric constraint upon the approximating posterior distribution.
Unfortunately, this assumption can often severely inhibit the quality of the approximate posterior should the true posterior not belong to the chosen family of probability distributions, as often happens with \glspl{GP} \citep{havasi_inference_2018}. 
The most common (asymptotically) exact inference method for \glspl{GP} is \gls{MCMC}. However, sampling can be problematic if the posterior distribution is non-convex as the sampler can become trapped in local modes \citep{rudoy_monte_2006}. Additionally, \gls{MCMC} does not enjoy the same computational scalability as \gls{VI}, and for this reason it is impractical for modelling problems with a large number of observations. 

In this work we propose the use of \gls{svgd} \citep{liu_stein_2016}, a non-parametric \gls{VI} approach, as an effective inference method for \gls{GP} models. 
\gls{svgd} can be thought of as a particle-based approach, whereby particles are sequentially transformed until they become samples from an arbitrary variational distribution that closely approximates the posterior of interest. \gls{svgd} can be considered a hybrid of \gls{VI} and Monte Carlo approaches, yielding benefits over both. The first benefit is removing the  parametric assumption used in \gls{VI}. The result of this is that inference through \gls{svgd} allows a richer variational distribution to be learned. 
A second benefit is that we do not need to compute the acceptance step required in \gls{MCMC}. This leads to greater efficiency in \gls{svgd} as we are only required to compute the score function, an operation that can be accelerated for big datasets using the subsampling trick in \secref{sec:svgp:outline}.
Finally, through the use of a kernel function acting over the set of particles, \gls{svgd} encourages full exploration over the posterior space, meaning that we are able to better represent the uncertainty in multimodal posteriors. This is a critical difference in the quality of inference that is possible through \gls{svgd} in comparison to alternative methods and we provide compelling empirical evidence to support this in \secref{sec:exps:multimodal}. \tabref{tab:compativeMethods} shows the position of this work within the current literature. 

Our article demonstrates how to use \gls{svgd} to fit \glspl{GP} to both Gaussian and non-Gaussian data, including when computational scalability is addressed through an inducing point representation of the original data. We prove that the \gls{svgd} scheme reduces the \gls{kl} divergence to the target distribution on each iteration. We empirically demonstrate the performance of \gls{svgd} in a range of both classification and regression datasets, comparing against traditional \gls{VI} and modern implementations of \gls{MCMC} for \glspl{GP}, including in a large-scale spatiotemporal model for air quality in the UK. We release, at {\tt https://gitbhub.com/RedactedForReview},
code for reproducing the experiments in \secref{sec:exps}, %
and a general library for fitting GPs using \gls{svgd} based entirely upon GPFlow \citep{matthews_gpflow:_2017} and TensorFlow \citep{abadi_tensorflow_2016}. For a demonstration of package see \appref{sec:app:demo}.

\begingroup
\begin{table*}[t]
\centering
\caption[caption]{Features of key inference methods for \gls{GP} models.}
\label{tab:compativeMethods}
\resizebox{\textwidth}{!}{%
\setlength{\tabcolsep}{3pt}
\begin{tabular}{lccccc}
\hline
Reference & $p(\by | \f)$  & Sparse                & Approx.\ posterior          & Hyperparams & Inference\\
\hline
\cite{opper_variational_2008}         & Binary   &  \xmark & Gaussian & Point estimate & Variational  \\
\cite{titsias_variational_2009}          & Gaussian       & \cmark & Gaussian & Point estimate & Variational  \\
\cite{nguyen_automated_2014}          & Any & \xmark & Gaussian mixture   & Point estimate  & Variational \\
 \cite{hensman_mcmc_2015}          & Any & \cmark & True posterior          & Marginalised & MCMC      \\
This work & Any & \cmark & True posterior & Marginalised & \gls{svgd}
 
 \\
\hline
\end{tabular}%
}
\end{table*}
\endgroup

\section{Stein Variational Gradient Descent}\label{sec:background:svgd}
\paragraph{Stein's discrepancy} The foundation for \gls{svgd} is Stein's identity \citep{stein_bound_1972}. For an arbitrary (continuously differentiable) density of interest $p$, 
\begin{align}
\label{equn:SteinIdentity}
    \mathbb{E}_{\bparticle\sim p}\Big[\underbrace{\phi(\bparticle) \nabla_{\bparticle}\log p(\bparticle)  + \nabla_{\bparticle}\phi(\bparticle)}_{\text{Stein operator: }\mathcal{A}_p\phi(\bparticle)}\Big]=0,
\end{align}
where $\phi$ belongs to a family of smooth functions $\mathcal{F}$. Stein's identity has found uses in a range of modern machine learning problems including variance reduction \citep{oates_control_2017}, model selection \citep{kanagawa_kernel_2019} and generative modelling \citep{pu_vae_2017}. In our application, $p$ will be the posterior density of interest. We can identify the internals of the expectation in \eqref{equn:SteinIdentity} as the \textit{Stein operator}, a quantity we denote as $\mathcal{A}_{p}\phi(\bparticle)$.

Stein's identity \eqref{equn:SteinIdentity} can be used to give  a notion of \textit{distance} between any two probability distributions. Replacing the density $p$ under which we evaluate the expectation in \eqref{equn:SteinIdentity} with a second density $q$, \eqref{equn:SteinIdentity} is 0 if and only if $p = q$. For a class of functions $\mathcal{F}$, the Stein discrepancy is defined to be
\begin{align}
    \label{equn:SteinDisc}
    \sqrt{\mathbb{D}(q , p)} = \sup_{\phi \in \mathcal{F}} \mathbb{E}_{\bparticle\sim q}\left[\trace(\mathcal{A}_{p}\phi(\bparticle)) \right].
\end{align}
Note that when $\phi$ is $d\times 1$ vector-valued function, the expectation term in \eqref{equn:SteinIdentity} yields a $d \times d$ matrix so the trace operator is applied (i.e., $\mathbb{E}\left[ \operatorname{trace}(\mathcal{A}_p\phi(\lambda))\right]$) to conserve the computation of a scalar value.

Taking $\mathcal{F}$ to be the unit ball of the \gls{rkhs} $\mathcal{H}^d$ of a positive definite kernel $\kappa(x, x')$,
we can optimise \eqref{equn:SteinDisc} explicitly \citep{liu_kernelized_2016}.
This functional optimisation yields the following closed form solution to \eqref{equn:SteinDisc}
\begin{align}
    \label{equn:BetaDef}
    \hat{\phi}(\bparticle) = \beta(\bparticle)/\norm{\beta}_{\mathcal{H}^d} \ \mbox{where} \ 
    \beta(\bparticle) = \mathbb{E}_{\bparticle'\sim q}\left[\mathcal{A}_p \kappa(\bparticle, \bparticle')\right].
\end{align}

\paragraph{Stein Variational Gradient Descent} 
In \gls{svgd}, as in classical \gls{VI}, we approximate the true posterior distribution $p$ with a variational distribution $q$ that minimises the \gls{kl} divergence between $p$ and $q$.
The innovation in \gls{svgd} is that we assume no parametric form for $q$.  From an arbitrary initial distribution $q_0$, \gls{svgd} iterates through a series of pushforward transformations that reduce the \gls{kl} divergence between the target distribution and the distribution $q_t$ after $t$ iterations. 

In particular, the transformation is defined by considering a mapping $\mathcal{T}(\bparticle) = \bparticle+\epsilon\phi(\bparticle)$ for an arbitrary function $\phi$ and perturbation magnitude $\epsilon$, where $\bparticle\sim q_t$. The transformed distribution $q_{t+1}$ is the distribution of $\mathcal{T}(\bparticle)$. Following \cite{gorham_measuring_2017} and \cite{liu_kernelized_2016}, we assume $\phi$ lives in the \gls{rkhs} $\mathcal{H}^d$. 
Under this assumption, 
it can be shown that
\begin{align}
    \label{equn:SVGD}
    \left.\nabla_{\epsilon} \operatorname{KL}\left(
    q_{t+1}
    \| p\right)\right|_{\epsilon=0}&=-\mathbb{E}_{\bparticle\sim q_t%
    }\left[\operatorname{trace}\left(\mathcal{A}_{p} \boldsymbol{\phi}(\bparticle)
    \right)\right].
\end{align}
Comparing \eqref{equn:SVGD} and \eqref{equn:SteinDisc} we see that using $\phi=\hat{\phi}$ from \eqref{equn:BetaDef} maximally decreases the \gls{kl} divergence. 

To implement the recursion, we maintain a finite set of $J$ samples that empirically represent $q$,  referred to as \textit{particles}. These particles $\Particle = \{\bparticle^{j}\}_{j=1}^{J}$ are initially sampled independently from $q_0$, which is typically taken to be the prior distribution.\footnote{In the asymptotic limit, the final particle values are invariant to the initial distribution that particles are initialised from \citep{papamakarios_normalizing_2019}} The transformation $\mathcal{T}$ is then applied repeatedly to the set of particles, where at each stage the optimal $\phi$ from \eqref{equn:BetaDef} is estimated empirically using the particles $\Particle_t = \{\bparticle^m_t\}_{j=1}^J$ at the $t^{\text{th}}$ iteration: 
\begin{align}
    \label{equn:SVGD_update}
    \hat{\phi}_{\Lambda_t}(\bparticle) = \frac{1}{J}\sum^J_{j=1}\bigg[\underbrace{\kappa(\bparticle^{j}_{t}, \bparticle)\nabla_{\bparticle}%
    \log p(\bparticle_{t}^{j})}_{\text{Attraction}} + \underbrace{\nabla_{\bparticle}%
    \kappa(\bparticle_{t}^{j}, \bparticle)}_{\text{Repulsion}}  \bigg].
\end{align}
When the process terminates after $T$ iterations, each of the particles is a sample from a distribution $q_T$ with low \gls{kl} divergence from the target $p$, and we can use the particles in the same way as a standard Monte Carlo sample.

Examining the update step in \eqref{equn:SVGD_update}, it can be seen that the first term transports particles towards areas in the posterior distribution that represent high probability mass. Conversely, the second term is the derivative of the kernel function; a term that will penalise particles being too close to one another (see \appref{sec:app:repulsion}). In the case that $J=1$, the summation in \eqref{equn:SVGD_update} disappears, and the entire scheme reduces to regular gradient based optimisation. Additionally, there is no danger of running \gls{svgd} with $J$ too large as, by the \textit{propagation of chaos} \citep{kac_probability_1976}, the final distribution of the $i^{\text{th}}$ particle is invariant to $J$ as the number of iteration steps $T\rightarrow\infty$ \citep{liu_stein_2016}.

\paragraph{Connection to variational inference}
Typically, in \gls{VI} we minimise the \gls{kl} divergence between a $\varParam$-parameterised variational distribution $q_{\varParam}(\bparticle)$ and the target density:
\begin{align}
\label{equn:VI_KL}
    \varParam^{\star} & = \argmin_{\varParam}\KL(q_{\varParam}(\bparticle)||p(\bparticle)).
\end{align}
$\varParam$ often parameterises a family $\{q_\varParam\}$ of Gaussian distributions. The resultant parameters $\varParam^{\star}$ are then used to form the optimal variational distribution $q^{\star}_{\varParam}(\bparticle)$, used in place of the intractable $p(\bparticle)$. 

In a regular \gls{VI} framework, the explicit form placed on $q$ can be highly restrictive, particularly if the true posterior density is not well approximated by the variational family selected. The nonparametric approach of \gls{svgd} allows for a more flexible representation of the posterior geometry beyond the commonly used Gaussian distribution used in regular \gls{VI}. An additional advantage is that \gls{svgd} only requires evaluation of the posterior's score function, a quantity that is invariant to the normalisation constant and can be unbiasedly approximated in large data settings.

\paragraph{Related \gls{svgd} work} \gls{svgd} has been used in the context of fitting Bayesian logistic regression and Bayesian neural networks \citep{liu_stein_2017}. Further, \gls{svgd} was used in the context of variational autoencoders (VAE) to model the latent space \citep{pu_vae_2017}. By relaxing the Gaussian assumption that is typically made of the latent space, it was possible to learn a more complex distribution over the latent space of the VAE. Further examples of the applications of \gls{svgd} can be found in reinforcement learning \citep{liu_stein_2017},  Bayesian optimisation \citep{gong_quantile_2019}, and in conjunction with deep learning \citep{grathwohl_cutting_2020}. Theoretical analysis has also established connections between \gls{svgd} and the overdamped Langevin diffusion \citep{duncan_geometry_2019}, and black-box variational inference \citep{chu_equivalence_2020}.

This article leverages the effective and efficient  \gls{svgd} optimisation framework to address the computational and multi-modality challenges endemic in %
\gls{GP} inference. We show that compared to standard inference approaches for GPs, the \gls{svgd} framework offers the best trade-off between accuracy, computational efficiency and model flexibility. 

\begin{figure*}[ht]
\begin{minipage}{0.49\textwidth}
\begin{algorithm}[H]
\caption{Pseudocode for fitting a Gaussian process using $T$ iterations of SVGD.}
\label{alg:SteinGP}
\begin{algorithmic}
\REQUIRE Base distribution $q_0$. Target distribution $p(\bparticle \given X,\by)$ where $\bparticle=\{\btheta,\boldsymbol{\nu}\}$.
\STATE Create $\Particle_0=\{\bparticle^j_0\}_{j=1}^J$ where $\bparticle_0^j\stackrel{\rm iid}{\sim} q_0.$
\FOR{t in 1:T} 
    \FOR{j in 1:J}
    \STATE $\bparticle^{j}_{t} \leftarrow \bparticle^j_{t-1} + \epsilon\hat{\phi}_{\Particle_{t-1}}(\bparticle^j_{t-1}) \,\,$ (see \eqref{equn:SVGD_update})
  \ENDFOR
    \STATE $\Particle_{t}=\{\bparticle_{t}^j\}_{j=1}^J$
\ENDFOR
\RETURN $\Particle_T$
\end{algorithmic}
\end{algorithm}
\end{minipage}
\hfill
\begin{minipage}{0.49\textwidth}
\begin{algorithm}[H]
\caption{Pseudocode for predictive inference over test inputs $X^{\star}$.}
\label{alg:SteinGPPredict}
\begin{algorithmic}
\REQUIRE Learned set of particles $\{\bparticle_j\}_{j=1}^J$
\STATE Initialise {\tt sample}$=\{\}$
\FOR{j in 1:J}
        \STATE Set $\left(\btheta, \boldsymbol{\nu}\right)$ = $\bparticle^j $
    \FOR{k in 1:K}
        \STATE Sample $\by^*\sim p(\cdot\,|\,X^*,\btheta,\boldsymbol{\nu})$
        \STATE Append $\by^*$ to {\tt sample}
    \ENDFOR
\ENDFOR
\RETURN 
{\tt mean(sample), var(sample)}
\end{algorithmic}
\end{algorithm}
\end{minipage}
\end{figure*}

\section{Gaussian Processes}\label{sec:svgg:GP}
Consider data $(X, \f, \by)=\{\x_i, f_i, y_i\}^N_{i=1}$ where $x_i\in\mathbb{R}^d$, and $y_i \in \mathbb{R}$ is a stochastic observation depending on $f_i=f(x_i)$ for some latent function $f$. Let $k_{\btheta}(\x, \x')$ be a positive definite kernel function parameterised by a set of hyperparameters $\btheta$ with resultant Gram matrix $\Kxx=k_{\btheta}(\x, \x')$. Following standard practice in the literature and assuming a zero mean function, we can posit the hierarchical \gls{GP} framework as 
\begin{align}
\label{equn:GPModel}
     p(\by \given \f , \btheta)& = \prod\limits_{i=1}^Np(y_i \given f_i, \btheta), \nonumber \\  \f\given X, \btheta &\sim \mathcal{N}(0, \Kxx), \\ \btheta &\sim p_0. \nonumber
\end{align}

From the generative model in \eqref{equn:GPModel}, we can see that the posterior distribution of a \gls{GP} is
\begin{equation}
    \label{equn:GPPosterior}
    p(\f, \btheta \given \by) = \frac{1}{C}p(\by \given \f,\btheta) p(\f \given \btheta)p_0(\btheta),
\end{equation}
where $C$ denotes the unknown normalisation constant of the posterior. Often we are interested in using the posterior to make new function predictions $f^{\star}$ for test data $X^{\star}$, 
\begin{equation}
    \label{equn:GPPredictive}
    p\left(\mathbf{f}^{\star} | \mathbf{y}\right)=\iint p\left(\mathbf{f}^{\star} | \mathbf{f}, \boldsymbol{\theta}\right) p(\mathbf{f}, \btheta | \mathbf{y}) \,\mathrm{d} \btheta \,\mathrm{d} \mathbf{f}.
\end{equation}
When the likelihood $p(y_i\,|\,f_i,\btheta)$ is Gaussian, the posterior predictive distribution conditional on $\btheta$ is analytically available as we can marginalise $\f$ out from \eqref{equn:GPPredictive}, and inference methods focus on $\btheta$.
For non-Gaussian likelihoods, we must approximate the integral over $\by$ using alternative approaches. Some of the most common methods are Laplace approximations \citep{williams_bayesian_1998}, expectation-propagation \citep{minka_family_2001, hernandez-lobato_scalable_2015}, \gls{MCMC} \citep{murray_elliptical_2010}, or \gls{VI} \citep{opper_variational_2008}. 

Even when $\f$ can be marginalised, computing \eqref{equn:GPPredictive} requires the inversion of $\Kxx$, an operation that requires $\mathcal{O}(N^3)$ computation. To reduce this cost when $N$ is large, one can introduce a set of \textit{inducing points} $Z = \{\mathbf{z}_{i}\}_{i=1}^M$ that live in the same space as $X$, such that $M \ll N$ \citep{snelson_sparse_2006}. The fundamental assumption made in the majority of sparse frameworks \citep[see][]{quinonero-candela_unifying_2005} is that the elements of $\f$ and $\f^{\star}$ are conditionally independent given $\bu=f(Z)$. Inference within this model only requires inversion of the Gram matrix $\Kzz = k_{\btheta}(Z, Z)$, which reduces the cost for posterior inference from $\mathcal{O}(N^3)$ to $\mathcal{O}(NM^2)$.

Inference for sparse GP models has thus far either required the GP hyperparameters $\btheta$ to be fixed so that \gls{VI} can be deployed \citep[e.g.][]{titsias_variational_2009, hensman_gaussian_2013}, or has used extremely computationally-intensive \gls{MCMC} schemes \citep[e.g.][]{hensman_mcmc_2015}. A full review of scalable \glspl{GP} can be found in \citet{liu_when_2019}. 

We demonstrate that \gls{svgd} is able to retain the ability to carry out joint inference over $\f$, $\btheta$ and, where necessary, $\bu$ without incurring the large overheads of MCMC schemes. In the rest of this article we do not explicitly discuss inference in sparse \glspl{GP}, but all our results apply equally to sparse formulations as to the full \gls{GP} models discussed; the sparse approximation simply corresponds to a variant \gls{GP} \citep{quinonero-candela_unifying_2005}.

\begin{table*}[ht]
\centering
\caption{Mean test log-likelihoods (larger is better) over 5 independent data splits, with bold values indicating the best performing method. Our SteinGP with 2, 5, 10 and 20 particles is compared against a \gls{GP} fitted using \gls{VI}, maximum likelihood (ML) and Hamiltonian Monte-Carlo (HMC) (Note, HMC is used for classification datasets (listed in blue) as an ML approach is intractable). For brevity, only the datasets where there is a significant difference between the best performing and one, or more, alternative methods are reported here. The full table can be found in \tabref{tab:fullUCITestLL} in \appref{app:fullUCILL}}
\resizebox{\textwidth}{!}{
\begin{tabular}{lcccccc}
  \hline
Dataset & SteinGP2 & SteinGP5 & SteinGP10 & SteinGP20 & VI & ML/HMC \\ 
  \hline
  Airfoil & $\mathbf{0.06\pm 0.04}$ & $ 0.06  \pm 0.04 $ & $ 0.05  \pm 0.06 $ & $ 0.05  \pm 0.05 $ & $ 0.03  \pm 0.03 $ & $ 0.03  \pm 0.03 $ \\ 
  \clf{Blood} & $ -0.6  \pm 0.05 $ & $ -0.6  \pm 0.04 $ & $ -0.6  \pm 0.05 $ & $ -0.61  \pm 0.04 $ & $\mathbf{-0.51\pm 0.05}$ & $ -0.52  \pm 0.06 $ \\ 
  \clf{Breast Cancer} &  $-0.08 \pm 0.04$ &  $-0.08 \pm 0.02$ & $-0.08 \pm 0.02$ & $\mathbf{-0.08 \pm 0.01}$ & $-0.65 \pm 0.09$ & $\mathbf{-0.08 \pm 0.04}$ \\
  Challenger & $ -1.53  \pm 0.45 $  & $ -1.52  \pm 0.43 $ & $\mathbf{-1.46  \pm 0.32}$  & $-1.53  \pm 0.41$   & $ -1.51  \pm 0.3 $ & $-1.51\pm 0.3$ \\ 
  Concreteslump & $1.08\pm 0.39$ & $ 1.07  \pm 0.41 $ & $ \mathbf{1.06  \pm 0.4} $ & $ 1.08  \pm 0.39 $ & $ 0.13  \pm 1.14 $ & $ 0.13  \pm 1.14 $ \\ 
  \clf{Fertility} &  $-0.44 \pm 0.03$ &  $-0.44 \pm 0.03$ & $-0.43 \pm 0.02$ & $\mathbf{-0.42 \pm 0.02}$ & $-0.70 \pm 0.08$ & $-0.54 \pm 0.02$ \\
  Gas & $ 0.88  \pm 0.11 $ & $ 0.88  \pm 0.11 $ & $\mathbf{0.89\pm 0.1}$ & $ 0.88  \pm 0.11 $ & $ 0.79  \pm 0.11 $ & $ 0.79  \pm 0.11 $ \\ 
  \clf{Hepatitis} & $ -0.41  \pm 0.07 $ & $ -0.41  \pm 0.07 $ & $ -0.42  \pm 0.07 $ & $\mathbf{-0.4\pm 0.07}$ & $ -0.69  \pm 0 $ & $ -0.44  \pm 0.04 $ \\ 
  Parkinsons & $ 4.12  \pm 0.05 $ & $ 4.12  \pm 0.05 $ & $\mathbf{4.14\pm 0.03}$ & $ 4.13  \pm 0.06 $ & $ 3.95  \pm 0.04 $ & $ 3.95  \pm 0.04 $ \\ 
  \clf{Spectf} & $-0.26\pm 0.01$ & $ -0.26  \pm 0.01 $ & $ -0.26  \pm 0.01 $ & $ \mathbf{-0.26  \pm 0.01} $ & $ -0.69  \pm 0 $ & $ -0.68  \pm 0.03 $ \\ 
  Winewhite & $ 0.56  \pm 0.05 $ & $ 0.57  \pm 0.05 $ & $\mathbf{0.57\pm 0.05}$ & $ 0.57  \pm 0.05 $ & $ 0.49  \pm 0.05 $ & $ 0.55  \pm 0.05 $ \\ 
   \hline
\end{tabular}
}
\label{tab:test_ll_sig}
\end{table*}

\section{Stein Variational Gaussian Processes}\label{sec:svgp:outline}
\paragraph{Gaussian data} Recalling the posterior distribution \eqref{equn:GPPosterior} of a \gls{GP}, we seek to approximate this distribution using a finite set of particles learned through \gls{svgd}. When the data likelihood is Gaussian, we are able to analytically integrate out $f$ and use \gls{svgd} to learn the kernel hyperparameters and observation noise $\sigma^2$, which comprise $\btheta$. This means that each particle $\bparticle^j$ represents a sampled $\btheta$ value. \gls{svgd} is useful even in these cases, as estimating kernel parameters such as the lengthscale can be particularly challenging for \gls{VI} and \gls{MCMC}. This is due to their unidentifiable nature that often manifests itself through a multimodal marginal posterior distribution (see \secref{sec:exps:multimodal} for an example of this). 
Through a set of interacting particles, \gls{svgd} is able to efficiently capture these modes. Accounting for posterior mass beyond the dominant mode is of utmost importance when trying to give realistic posterior predictions \citep{palacios_non-gaussian_2006, gelfand_handbook_2010}.

\paragraph{Non-Gaussian data} In the general \gls{GP} setup in \eqref{equn:GPModel}, where $y_i \given f_i, \btheta$ is non-Gaussian, we are required to learn the latent values $\f$ of the \gls{GP} in addition to $\btheta$. To decouple the strong dependency that exists between $\f$ and $\btheta$, we centre (or `whiten') the \gls{GP}'s covariance matrix such that $\f=L_{\btheta}\boldsymbol{\nu}$, where $L_{\btheta}$ is the lower Cholesky decomposition of the Gram matrix $\Kxx$, and $\nu_i\sim\mathcal{N}(0, 1)$. Applying such a transformation has been shown to enhance the performance of inferential schemes in the \gls{GP} setting \citep{murray_elliptical_2010, hensman_mcmc_2015, salimbeni_doubly_2017}. Once whitened, we can use the joint posterior distribution $p(\btheta, \boldsymbol{\nu} \given X, \by)$ as the target distribution for \gls{svgd} and \textit{post hoc} deterministically transform the posterior samples to give $p(\btheta, \f \given X, \by)$.

\gls{svgd} requires evaluation of the score function of the density at the current particle values to evaluate \eqref{equn:SVGD_update}. Using automatic differentiation, this  is a trivial task. However, it is accompanied by a computational cost that scales quadratically with $N$ since we can no longer marginalise $\f$. For this reason, in datasets surpassing several thousand datapoints, we estimate the score function using subsampled mini-batches $\Psi\subset \{1, \ldots , N\}$. The end result of this is a score function approximation that can be written as
\begin{align}
    \label{equn:batched_score}
    \nabla_{\btheta, \boldsymbol{\nu}} \log p(\btheta, \boldsymbol{\nu} \given X, \by) \approx & \nabla_{\btheta, \boldsymbol{\nu}}\log p_{0}(\btheta, \boldsymbol{\nu}) + \nonumber\\ & \frac{N}{|\Psi|}\sum_{i\in \Psi}\nabla_{\btheta, \boldsymbol{\nu}}\log p(\by_{i} \given \btheta, \nu_i).
\end{align}

\paragraph{Posterior predictions} Once a set of particles $\Particle$ has been learned, we can use the value of each particle to make predictive inference. Recalling that the primary motivation of \gls{svgd} is to enable accurate predictive inference in posterior distributions with complex and multimodal geometries, naively taking the mean particle value for each parameter as the final estimate of each parameter in the \gls{GP} could become problematic when the posterior is complex. For this reason, to obtain a predictive mean and variance, we sample $K$ times from the predictive posterior of the \gls{GP} for each of the $J$ particles and compute the mean and variance of the predictive samples. To elucidate this notion, the procedure is summarised in \algref{alg:SteinGPPredict} .

\paragraph{Optimisation guarantees}
We would like to show that the SteinGP in Algorithm \ref{alg:SteinGP} iteratively improves the posterior approximation. Consider the variational distributions $q_t$ of the particles at time $t$. We can show that the Kullback-Leibler divergence between these distributions and the target $p$ decreases monotonically as $t$ increases. The following result is similar to that of \cite{liu_stein_2017} and \cite{korba_non-asymptotic_2020} but with different assumptions and an alternative proof construction. This serves as a useful starting point for more rigorous analysis of \glspl{GP} (e.g., \citep{burt_convergence_2020}).

\begin{theorem} Consider \gls{svgd} in a general model with $\log p(\bparticle)$ at least twice continuously differentiable and $\nabla\log p(\bparticle)$ is smooth with Lipschitz constant $L$. The Kullback-Leibler divergence between the target distribution $p$ and its SVGD approximation $q_t$ at iteration $t$ is monotonically decreasing, satisfying
\begin{align}
    \KL(q_{t+1} || p) - & \KL (q_{t}|| p)  \leq -\epsilon \mathbb{D}(q_t , p)^2 (1- \nonumber\\ & \epsilon\mathbb{E}_{\bparticle\sim q_t}\left[ L\kappa(\bparticle, \bparticle)/2 + \nabla_{\bparticle, \bparticle}\kappa(\bparticle, \bparticle) \right])
\end{align}
\label{thm}
\end{theorem}

\begin{figure}[ht]
    \centering
    \includegraphics[width=0.8\textwidth, height=0.25\textheight]{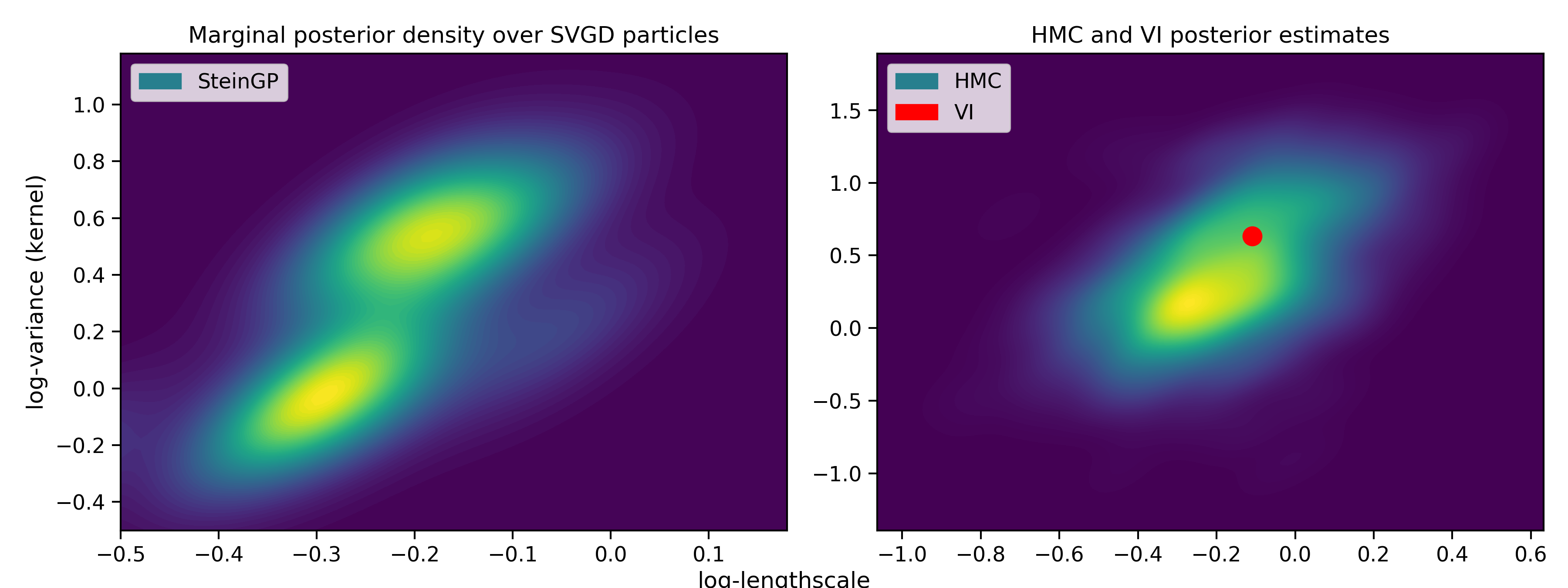}
    \caption{The marginal posterior distributions of the \gls{GP} model's kernel parameters in the multimodal example of \secref{sec:exps:multimodal}. The left panel shows the bimodal posterior learned through SVGD, whereas the HMC inferred posterior appears to have just explored a single mode. In order to obtain a tractable \gls{elbo} objective for the VI optimiser, model hyperparameters are treated as variational parameters \citep{titsias_variational_2009}. Consequently, only scalar values for the hyperparameters can be learned. This scalar pair is depicted by the red point in the right panel.}
    \label{fig:multimodalPosteriors}
\end{figure}

It can be shown that for the \gls{GP} models considered in this paper, the gradients are Lipschitz smooth and therefore Theorem \ref{thm} holds. Additionally, Theorem 8 of \cite{gorham_measuring_2017} establishes weak convergence for a sequence of probability measures $(q_t)_{t \geq 1}$, where $q_t \implies p$ if $\mathbb{D}(q_t, p) \rightarrow 0$, and so it follows from Theorem \ref{thm}, that $q_t \implies p$ as $t \rightarrow \infty$.

\section{Experiments}\label{sec:exps}
\paragraph{Benchmark datasets} For the datasets used in \secref{sec:exps:UCI}, models are fitted using 70\% of the full dataset, whilst the remaining 30\% is used for model assessment. The partitioning of data, hyperparameter initialisation and computing environment used is the same for all models. For each dataset we standardise the inputs and outputs to zero mean and unit standard deviation.

\paragraph{\gls{svgd} implementation details} Throughout our experiments we use an RBF kernel $\kappa$ to specify the \gls{svgd} step \eqref{equn:SVGD_update} (which is an independent choice from the kernel within the \gls{GP} model). The RBF kernel's variance is set to 1, and we select the lengthscale at each iteration of \gls{svgd} using the median rule as in \cite{liu_stein_2016}. We run experiments with $J=2$, 5, 10 and 20 particles (labelled SteinGP2 through to SteinGP20).

Further experimental details are given in \appref{sec:app:expDetails}.

\subsection{UCI Datasets}\label{sec:exps:UCI}

We analyse 19 UCI datasets where the target is 1-dimensional. Datasets range in size from 23 to 5875 datapoints. As highlighted in \appref{app:UCIDatasets}, the targets in 6 of the datasets are binary, whilst the remaining 14 datasets are continuous values. To accommodate this, we use a \gls{GP} with a Bernoulli likelihood for the classification tasks, and a Gaussian likelihood for regression. For regression, inference is made of the model's hyperparameters and for classification we also infer the latent values of the GP. 

When the likelihood is Gaussian, we compare our SteinGP against a \gls{GP} that is fit using maximum likelihood (ML) and another \gls{GP} fit using \gls{VI} (VI). For Bernoulli likelihoods, maximum likelihood estimation is not feasible so we instead use \gls{HMC} for inference along with \gls{VI}. In \tabref{tab:test_ll_sig} we report the test log-likelihoods where there was a significant difference between one, or more, of the models. A full table can be found in \tabref{tab:fullUCITestLL}.

From \tabref{tab:test_ll_sig}, it can be seen in all but two of the experiments that a GP inferred using \gls{svgd} significantly outperforms comparative methods. Furthermore, \tabref{tab:test_ll_sig} shows that increasing the number of particles for  \gls{svgd} leads to larger test log-likelihood values.

\begin{table}[h]
\centering
\caption{Relative average computational runtimes of comparative methods. Results are reported relative to a SteinGP with 2 particles i.e., a value of 2 would indicate that method was two times slower than SteinGP2.}
\label{tab:RuntimesAvg}
\begin{tabular}{lccc}
\hline
     & ML & VI & HMC     \\ \hline
    Relative wall time & 1.8 & 1.3 & 5.6
 \\ \hline
\end{tabular}
\end{table}

For two particles, a SteinGP is almost always faster than comparative methods. The only exceptions to this are when either the maximum likelihood approach converges very quickly and the difference is then of the order of seconds, or the variational approach for non-conjugate inference. 

Across all UCI datasets we report the average runtime of each comparative method and report timings relative to a SteinGP with 2 particles \tabref{tab:RuntimesAvg}. It is clear that a SteinGP is a computationally efficient model, particularly compared to a \gls{HMC} approach, and a SteinGP consistently produces optimal predictive results. A full table of computational runtimes broken out by dataset can be found in \appref{app:UCIRuntimes}.

\subsection{Multimodal Inference}\label{sec:exps:multimodal}
A multimodal posterior is often the result of a misspecified model. However, posterior multimodality can also occur due to corrupted data. To see this, we use the 1-dimensional example from \cite{neal_monte_1997} whereby data is generated according to
\begin{align*}
    y_i = 0.3 + 0.3x_i+0.5\sin(2.7x_i) + \frac{1.1}{1+x_i^2} + \epsilon_i
\end{align*}
where $x_i \sim \mathcal{N}(0, 1)$ and $\epsilon_i \sim \mathcal{N}(0, \sigma^2)$. We set $\sigma = 1$ with probability 0.05, and $\sigma = 0.1$  otherwise, thus inflating the variance of some data points and creating outliers. We simulate 200 points from this model, with the first 100 used for the fitting of the \gls{GP} and remaining 100 used for evaluation \appref{app:Multimodal}. 

The data are modelled using a \gls{GP} that is equipped with a squared exponential kernel $k(x, x') = \alpha^2\exp\left(-(x-x')^2/2\ell^2\right)$. We assume the observation noise follows a zero-mean Gaussian distribution with variance $\sigma^2$. We therefore wish to learn the posterior distribution  $p(\btheta)$ where $\btheta=\{\alpha, \ell, \sigma\}$.

\begin{table}[h]
\centering
\caption{Predictive metrics on 100 heldout datapoints from the multimodal example in \secref{sec:exps:multimodal}. }
\label{tab:multimodalResults}
\begin{tabular}{lccc}
\hline
                      & SteinGP         & HMC           & VI     \\ \hline
RMSE                  & \textbf{0.39}  & 0.43          & 0.44   \\
Log posterior density & \textbf{-24.33} & -24.38        & -27.31 \\
Runtime (seconds)     & 24.8            & 50.4 & \textbf{14.7}   \\ \hline
\end{tabular}
\end{table}

We compare the posteriors of a SteinGP with 20 particles against posteriors sampled using \gls{HMC} in \figref{fig:multimodalPosteriors}. The point estimates inferred using \gls{VI} are included for a full and faithful comparison. Accurately quantifying the posterior uncertainty is not only useful for parameter interpretation, but also leads to higher quality predictive inference (see \tabref{tab:multimodalResults}). From \tabref{tab:multimodalResults} it is clear that at the cost of slightly longer computational runtime than \gls{VI}, a SteinGP will yield significantly better predictive inference. This can be seen through the improved \gls{RMSE} of the SteinGP when compared against a \gls{VI} and \gls{HMC} inferred model.

\subsection{Fully Bayesian Sparse Gaussian Process}\label{sec:exps:AQ}

As a final experiment, we consider a spatiotemporal air quality dataset comprised of 550,134 datapoints. Data is recorded hourly from 237 sensors across the \gls{UK}.  The time period of interest is February 1$^{\text{st}}$ to April 30$^{\text{th}}$ 2020; the period in which the \gls{UK} entered a Covid-19 lockdown. Consequently, the spatiotemporal dynamics of air quality levels are chaotic and challenging to model as many individuals adopt lifestyle changes with huge ramifications towards their contribution to air pollution (see \figref{app:AQ:Temporal}). Historically, fully Bayesian inference with \gls{GP} models would be infeasible on such datasets due to the challenging scaling of \gls{MCMC} samplers. However, using the batched approach in \eqref{equn:batched_score} we are able to make full Bayesian inference tractable.

\begin{figure}[ht]
    \centering
    \includegraphics[width=0.41\textwidth, height=0.34\textheight]{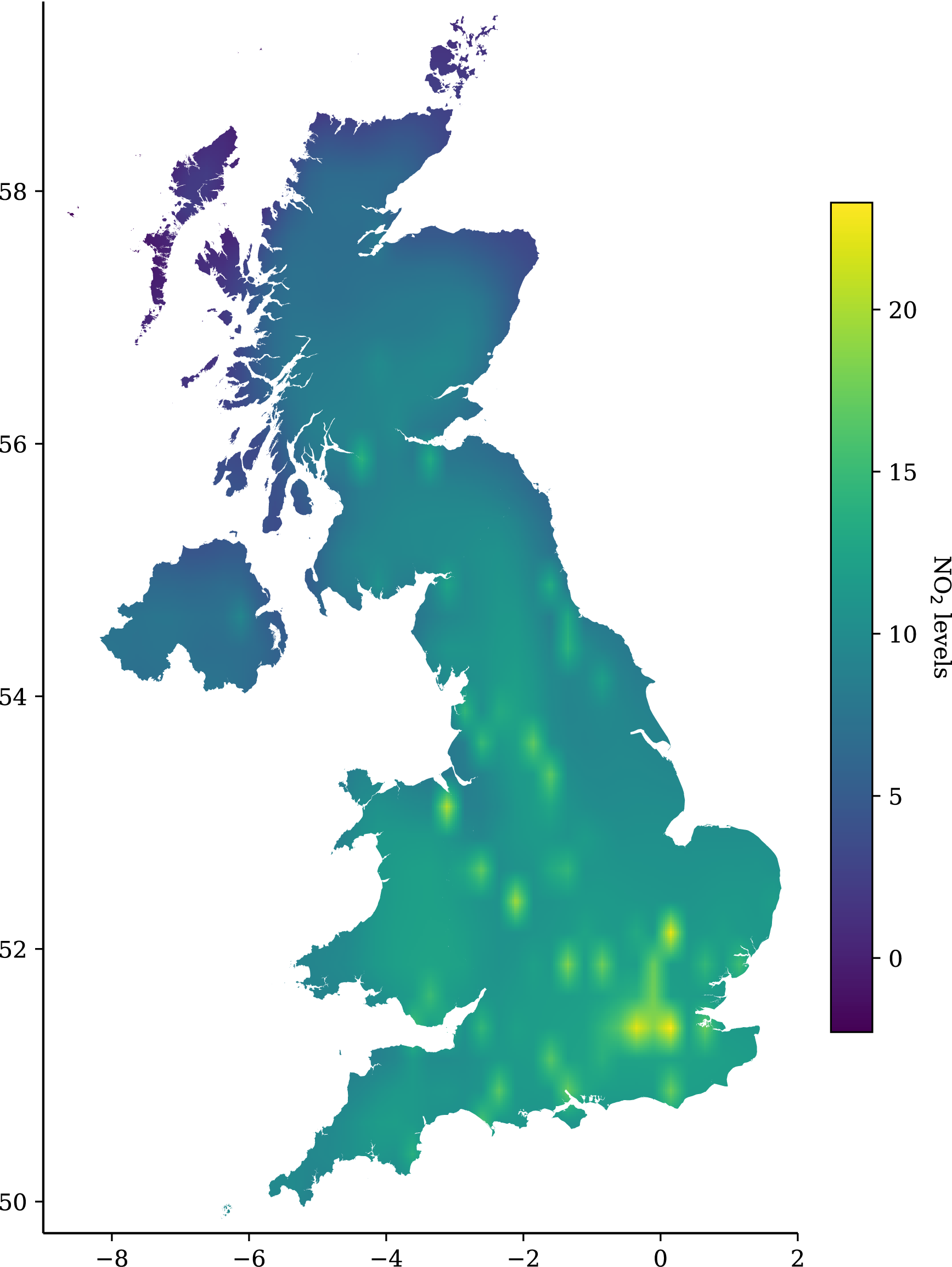}
    \caption{Inferred \notwo spatial surface ($\mu$gm$^{-3}$) in the \gls{UK} at 9AM on March 23$^{\text{rd}}$; the day that initial lockdown measures were announced.}
    \label{fig:lockdownSlice}
\end{figure}

To further demonstrate the efficacy of a SteinGP, we develop, in conjunction with climate scientists, a complex separable kernel that principally captures the complex spatiotemporal dynamics of atmospheric \gls{NO2}. Across the spatial dimensions we use a third-order Mat\'ern kernel and in the temporal dimension we use a product of a first-order Mat\'ern and a third-degree polynomial kernel to capture the temporal nonstationarity. We further include a white noise process. See \appref{app:kernelExpressions} for full expressions of these kernels and \appref{app:demo:AQ} for a demo implementation of this model.

\paragraph{Predictive performance}We assess the performance of our model by computing the predictive log-likelihood and the \gls{RMSE} of our model and compare against the state-of-art sparse \gls{GP} fitted using stochastic \gls{VI} \citep{titsias_variational_2009, hensman_gaussian_2013}. To induce sparsity, the same set 600 inducing points are used in both models. We initialise the inducing points through a \gls{DPP}, as per \cite{burt_convergence_2020}. 

\begin{table}[h]
\centering
\caption{Comparison of our SteinGP with 30 particles against a \gls{GP} fitted using stochastic \gls{VI} on the air quality data of \secref{sec:exps:AQ}. Standard errors are computed by fitting each model on 5 random splits of the data, with 30\% of the data being used for prediction.}
\label{tab:aqResults}
\begin{tabular}{lcc}
\hline
               & SteinGP   & SVI       \\ \hline
RMSE           & $0.82 \pm 0.09$ & $0.79 \pm 0.07$ \\
Log-likelihood & $-1.358 \pm 0.15$          &  $-1.342 \pm 0.11$        \\
\hline

\end{tabular}
\end{table}

We optimise both models by running the SteinGP for 1000 iterations and the stochastic VI model for 10000. For both models, we use a batch size of 256 and a learning rate of 0.001. 30 particles are used for the \gls{svgd} routine. Due to the size of the dataset, a single step of a \gls{HMC} sampler took over 10 minutes. Given that several thousand steps of the \gls{HMC} sampler would be required, a comparison against \gls{HMC} is infeasible.

\begin{figure}[ht]
    \centering
    \includegraphics[width=0.8\textwidth, height=0.22\textheight]{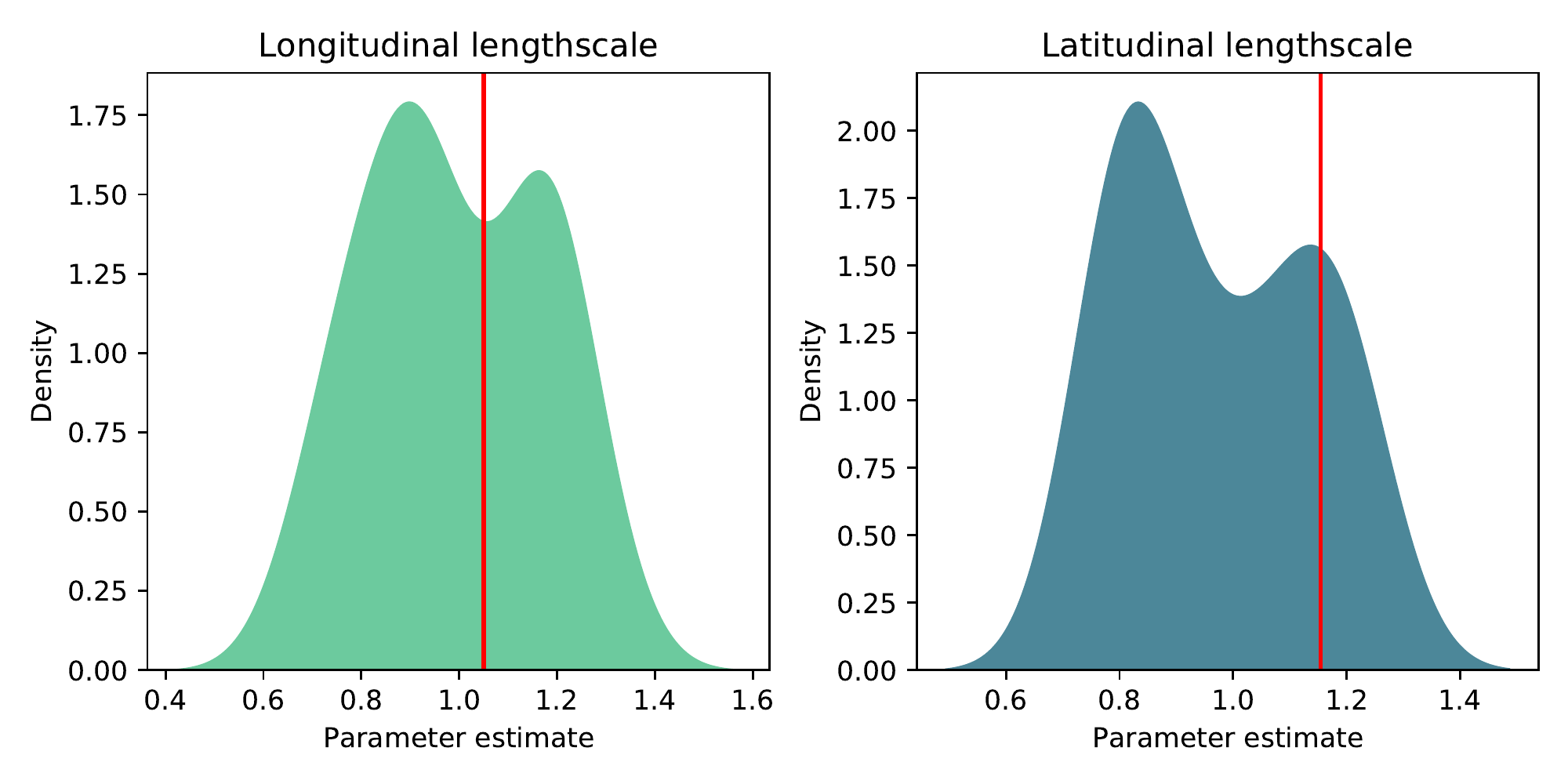}
    \caption{Spatial lengthscale posteriors for the air quality data in \secref{sec:exps:AQ}. The density for each lengthscale is estimated using the optimised particles in \gls{svgd} and the red line corresponds to the scalar estimates produced through a stochastic \gls{VI} scheme.}
    \label{fig:AQLengthscales}
\end{figure}

\tabref{tab:aqResults} shows there is no significant difference between the \gls{GP} fit using \gls{svgd} and stochastic \gls{VI}. Unsurprisingly, \gls{VI} is faster than \gls{svgd} with average computational wall times of 141 seconds and 332 seconds, respectively. However, this is to be expected due to the kernel computations that are required at each iteration of \gls{svgd}.%

\begin{figure}
\centering
\begin{subfigure}{.5\textwidth}
  \centering
  \includegraphics[width=0.95\linewidth]{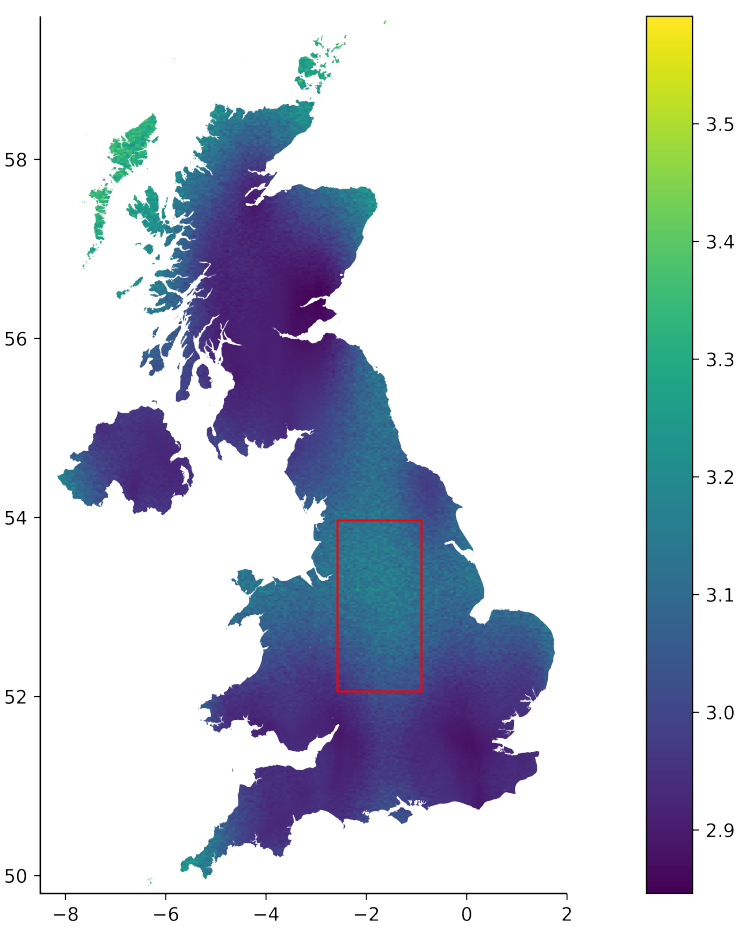}
  \caption{SteinGP inferred posterior uncertainty.}
    \label{fig:SteinGPPosteriorVariance}
\end{subfigure}%
\begin{subfigure}{.5\textwidth}
  \centering
  \includegraphics[width=0.95\textwidth]{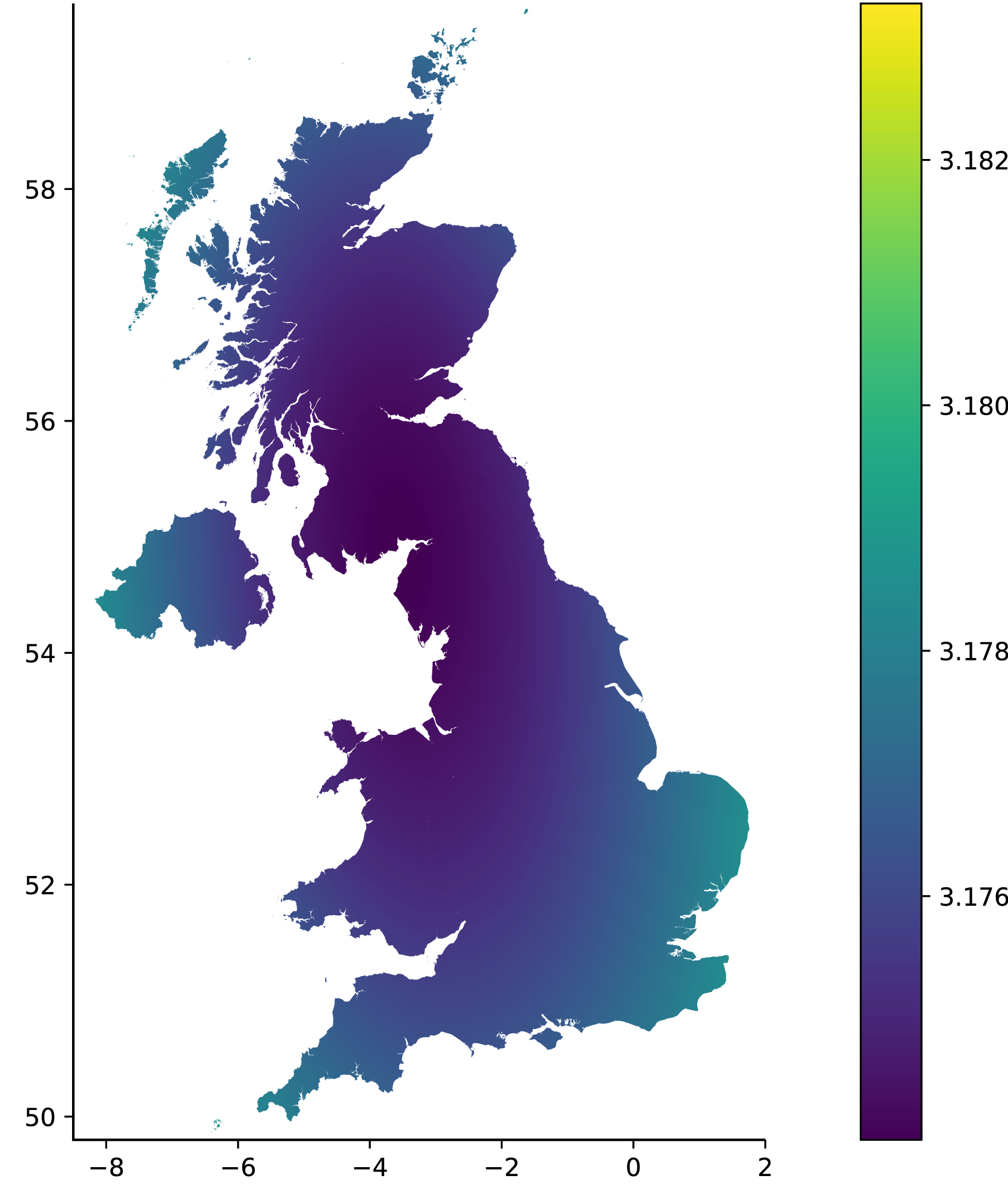}
  \caption{Stochastic \gls{VI} inferred posterior uncertainty.}
  \label{fig:SVGPPosteriorVariance}
\end{subfigure}
\caption{Posterior variances for a SteinGP (\ref{fig:SteinGPPosteriorVariance}) and stochastic \gls{VI} \gls{GP} (\ref{fig:SVGPPosteriorVariance}). Data within the red square in \ref{fig:SteinGPPosteriorVariance} was held back and predictions were then made across the entire UK. The lighter colours indicate a higher predictive variance; something that is expected when there are no observation present. We note the differing colour scales used in \ref{fig:SteinGPPosteriorVariance} and \ref{fig:SVGPPosteriorVariance}.}
\label{fig:test}
\end{figure}

\paragraph{Uncertainty quantification} Although the headline predictive performance of the two methods is comparable, the fully Bayesian treatment of the \gls{GP} model that SVGD enables leads to full posterior inference and improved uncertainty quantification. To see this, we hold out all stations in the midlands of the UK (see the red box in \figref{fig:SteinGPPosteriorVariance}) and re-fit both the SteinGP and the stochastic \gls{VI} counterpart. We then perform spatial interpolation over the midlands of the UK from February 1$^{\text{st}}$ through to April 30$^{\text{th}}$. In a model that is capable of generating effective predictive uncertainties, the predictive variance should be much larger over the midlands due to the lack of observations there.

Both models achieve comparable predictive metrics (\tabref{tab:aqSpatialResults}). However, the purpose of this study is to assess the predictive uncertainties that each model yields. It can be seen in \figref{fig:AQLengthscales} that estimating the spatial lengthscale parameters is challenging due to multimodalities. The SteinGP appears to have captured a secondary mode for each parameter. For completeness, we also give the scalar estimates produced by the stochastic VI procedure.

From \figref{fig:SteinGPPosteriorVariance} it can be seen that the richer posterior inference provided from a SteinGP results in more reasonable posterior uncertainty estimates than those displayed in \figref{fig:SVGPPosteriorVariance}. This can be seen by observing the increased uncertainty over the midlands of the UK. Noting the different legend colour scaling, it can be seen in comparison that the GP inferred using stochastic VI is incapable of capturing this behaviour and yields an almost uniform predictive variance across the UK.

\section{Discussion}
We have shown that \gls{svgd} can be used to provide comparable inferential quality to the gold standard \gls{HMC} sampler in \gls{GP} inference, while only requiring a computational cost comparable to \gls{VI}. The ability to carry out joint inference over latent function values and kernel hyperparameters allows for a full and proper consideration of uncertainty in the inference.

For simple problems where the true posterior is Gaussian, very few \gls{svgd} particles are required to achieve strong inference, as can be seen in the experiments carried out in \secref{sec:exps:UCI}. For more complex problems, such as those in \secref{sec:exps:multimodal}, a large number of particles can help capture subtle posterior geometries. 

Well quantified predictive uncertainties are critical when \gls{GP} models are being used in real-world scenarios. As shown in \secref{sec:exps:AQ}, a SteinGP is able to provide well quantified uncertainty estimates, even in big data scenarios. This, in conjunction with strong predictive performance, makes SteinGPs a useful tool for practitioners fitting \gls{GP} models.

\clearpage

\bibliographystyle{apalike}
\bibliography{main}
\clearpage

\begingroup

\section*{Supplementary Material}

\section{Particle repulsion}\label{sec:app:repulsion}
Reminding ourselves of the update step \eqref{equn:SVGD_update} of \gls{svgd}:
\begin{align*}
    \hat{\hat{\phi}}_{\Lambda_t}(\bparticle) = \frac{1}{J}\sum^J_{j=1}\left[\underbrace{\kappa(\bparticle_{t}^{j}, \bparticle)\nabla_{\bparticle} \log p(\bparticle_{t}^{j})}_{\text{Attraction}} + \underbrace{\nabla_{\bparticle}\kappa(\bparticle_{t}^{j}, \bparticle)}_{\text{Repulsion}}  \right],
\end{align*}
we demonstrate how particles are repelled from one another. If we take $\kappa(\cdot, \cdot)$ to be the radial basis function (a valid kernel when computing \gls{KSD} \citep{gorham_measuring_2017}), then we have
\begin{align}
    \label{equn:RBF}
    \kappa(\bparticle, \bparticle') = \exp\left(\frac{\norm{\bparticle - \bparticle'}^2}{-\ell^2}\right)
\end{align}
whereby
\begin{align}
    \label{equn:RBFDeriv}
    \nabla_{\bparticle}\kappa(\bparticle, \bparticle') & = -\frac{2 \bparticle-2 \bparticle'}{ \ell^{2}} \exp \left(\frac{\norm{\bparticle-\bparticle'}^{2}}{-\ell^{2}}\right) \nonumber \\
    & = -\frac{2 (\bparticle- \bparticle')}{\ell^{2}}\kappa(\bparticle, \bparticle').
\end{align}
Should particles be densely clustered, then the resultant Gram matrix will be dense. This will lead to a larger quantity being computed upon evaluation of \eqref{equn:RBFDeriv}, compared to when particles are sufficiently far from one another.

\section{Proof of Theorem 1}\label{sec:app:proof}

\gls{svgd} maps particles using $\bparticle_{t+1}={\mathcal T}(\bparticle_t)=\bparticle_t+\epsilon\hat{\phi}(\bparticle_t)$. Denote the corresponding mapping of densities by $q_{t+1}=T(q_t)$. We have that
\begin{align}
\KL(q_{t+1}||p) - \KL(q_{t}||p) & = \KL(T(q_{t})||p) - \KL(q_{t}||p) \nonumber\\
& = \KL(q_t ||T^{-1}(p))- \KL(q_{t}||p)\nonumber\\
&= \mathbb{E}_{\bparticle\sim q_t}\left[\log q_t(\bparticle) - \log T^{-1}(p)(\bparticle)\right] - \mathbb{E}_{\bparticle\sim q_t}\left[ \log q_t(\bparticle) - \log p(\bparticle)\right]\nonumber\\
&=\mathbb{E}_{\bparticle\sim q_t}\left[\log p(\bparticle) - \log T^{-1}(p)(\bparticle)\right].\label{equn:Proof1}
\end{align}

Under the  %
change of variable formula for densities, we have
\begin{align*}
    T^{-1}(p)(\bparticle) = p(\mathcal{T}(\bparticle))\cdot |\operatorname{det}(\nabla_{\bparticle}{\mathcal T}(\bparticle))|
\end{align*}
which allows us to rewrite \eqref{equn:Proof1} as 
\begin{align}
    \mathbb{E}_{\bparticle\sim q_t}\left[\log p(\bparticle) - %
    \log p(\bparticle+\epsilon\hat{\phi}(\bparticle))%
    -\operatorname{log}|\operatorname{det}(\nabla_{\bparticle}{\mathcal T}(\bparticle))|\right].
    \label{eq:toBound}
\end{align}

Assuming that $\nabla_{\bparticle} \log p(\bparticle)$ is Lipschitz smooth with constant $L$ and $\log p(\bparticle) \in C^2$, a second order Taylor series approximation of $\log p(\bparticle+\epsilon\hat{\phi}(\bparticle))$ about $\bparticle$ (assuming $\epsilon \ll 1$) %
lets us bound the first two terms in \eqref{eq:toBound} by 
\begin{align}
    \log p(\bparticle) - \log p(\bparticle + \epsilon\hat{\phi}(\bparticle)) &\leq -\epsilon\nabla_{\bparticle}\log p(\bparticle)^{\intercal}\hat{\phi}(\bparticle)+\frac{L\epsilon^2}{2}\hat{\phi}(\bparticle)^{\intercal}\hat{\phi}(\bparticle).%
    \label{eq:taylor}
\end{align}
Noting the definition of the Stein operator from \eqref{equn:SteinIdentity}, we have that $    -\epsilon\nabla_{\bparticle}\log p(\bparticle)^{\intercal}\hat{\phi}(\bparticle)=\operatorname{trace}(-\epsilon \mathcal{A}_{p}\hat{\phi}(\bparticle)-\epsilon\nabla_{\particle}\hat{\phi}(\bparticle))$. Note also that
${\mathcal T}(\bparticle) = \bparticle + \epsilon\hat{\phi}(\bparticle)$, and therefore $\nabla_{\bparticle}{\mathcal T}(\bparticle)=I+\epsilon\nabla_{\bparticle}\hat{\phi}(\bparticle)$. We can lower bound the final term in \eqref{eq:toBound} by first noting that by the approximate Neumann expansion of the inverse matrix $\nabla_{\bparticle}{\mathcal T}(\bparticle)^{-1}$,
\begin{align}
\nabla_{\bparticle}{\mathcal T}(\bparticle)^{-1} = (I+\epsilon\nabla_{\bparticle}\hat{\phi}(\bparticle))^{-1} \approx I - \epsilon\nabla_{\bparticle}\hat{\phi}(\bparticle) + (\epsilon\nabla_{\bparticle}\hat{\phi}(\bparticle))^2 
\label{eq:neumann}
\end{align}
which holds for $0<\epsilon\leq\rho(\nabla_{\bparticle}\hat{\phi}(\bparticle))^{-1}$, where $\rho(\cdot)$ is the matrix spectral norm. We can bound the final term in \eqref{eq:toBound} using the following lower bound, 
\begin{align}
\operatorname{log}|\operatorname{det}(\nabla_{\bparticle}{\mathcal T}(\bparticle))| \geq \sum_{i=1}^d(1-e_i^{-1}) = \trace (I-\nabla_{\bparticle}{\mathcal T}(\bparticle)^{-1})
\label{eq:logdet}
\end{align}
where $e_1,\ldots,e_d$ are the eigenvalues of $\nabla_{\bparticle}{\mathcal T}(\bparticle)$. Replacing the Neumann expansion for $\nabla_{\bparticle}{\mathcal T}(\bparticle)^{-1}$ in \eqref{eq:logdet}, gives the following lower bound,
\begin{align}
\operatorname{log}|\operatorname{det}(\nabla_{\bparticle}{\mathcal T}(\bparticle))| \geq
 \epsilon \nabla_{\bparticle} \cdot \hat{\phi}(\bparticle) - \epsilon^2 ||\nabla_{\bparticle}\hat{\phi}(\bparticle))||_{F}^2,     
 \label{eq:logdet2}
\end{align}
where $||\cdot||_F$ is the Frobenius norm. 

Combining \eqref{eq:taylor} and \eqref{eq:logdet2} gives the following upper bound for \eqref{eq:toBound},
    
\begin{align}
    \label{equn:ProofPartition}
    & \mathbb{E}_{\bparticle\sim q_t}\left[-\epsilon\nabla_{\bparticle}\log p(\bparticle)^{\intercal}\hat{\phi}(\bparticle)+\frac{L\epsilon^2}{2}\hat{\phi}(\bparticle)^{\intercal}\hat{\phi}(\bparticle) - \epsilon \nabla_{\bparticle} \cdot \hat{\phi}(\bparticle) + \epsilon^2 ||\nabla_{\bparticle}\hat{\phi}(\bparticle))||_{F}^2 \right]. \\
    = & \underbrace{-\epsilon\mathbb{E}_{\bparticle\sim q_t}\left[\mathcal{A}_{p}\hat{\phi}(\bparticle)\right]}_{B} + \underbrace{\mathbb{E}_{\bparticle\sim q_t}\left[\underbrace{\epsilon^2 ||\nabla_{\bparticle}\hat{\phi}(\bparticle))||_{F}^2}_{C1} +  \underbrace{\frac{L\epsilon^2}{2}\hat{\phi}(\bparticle)^{\intercal}\hat{\phi}(\bparticle)}_{C2}%
    \right]}_{C}.
\end{align}
By definition of the Stein discrepancy  \eqref{equn:SteinDisc}, $B=-\epsilon \mathbb{D}(q_t, p)^2$, and \eqref{equn:Proof1} becomes
\begin{align}
\label{equn:ProofProblem}
    \KL(q_{t+1}||p) - \KL(q_{t}||p) \leq -\epsilon \mathbb{D}(q_t, p)^2 + C.
\end{align}
Based on this, we must now show that $C$ is bounded, which we can do by considering each term individually.

C2: We can bound this term using the properties of the RKHS \citep{berlinet_reproducing_2004}. As $\hat{\phi} = (\hat{\phi}_1,\ldots,\hat{\phi}_d)'$ and $\hat{\phi}_i \in \mathcal{H}_0 \implies \hat{\phi} \in \mathcal{H}^d$ then 
\begin{align}
    \norm{\hat{\phi}(\bparticle)}^2_2 & = \sum\limits^d_{i=1}\hat{\phi}_i(\bparticle)^2 \nonumber \\
    & = \sum\limits^d_{i=1}\left( \langle\hat{\phi}_i(\cdot), \kappa(\bparticle, \cdot)\rangle_{\mathcal{H}_0}\right)^{2} \mbox{which follows from the RKHS properties}%
    \nonumber\\
    & \leq \sum\limits^d_{i=1}\norm{\hat{\phi}}_{\mathcal{H}_0}^2\norm{\kappa(\bparticle, \cdot)}^2_{\mathcal{H}_0} \mbox{by Cauchy-Schwarz} \nonumber \\
    & = \norm{\hat{\phi}}_{\mathcal{H}^d}^{2}\kappa(\bparticle, \bparticle') \nonumber\\
    & = \mathbb{D}(q_t, p)^2 \kappa(\bparticle, \bparticle') \mbox{ which follows by \eqref{equn:BetaDef}}\label{proof:CauchySchwarz}
\end{align}

C1: %
We upper bound the matrix norm $\epsilon||\nabla_{\bparticle}\hat{\phi}(\bparticle)||^2_F$ using the same RKHS property used in C2.

\begin{align*}
    \norm{\nabla_{\bparticle}\hat{\phi}(\bparticle)}^2_F & = \sum_{i=1}^d\sum_{j=1}^d \left(\frac{\partial \hat{\phi}_i(\bparticle)}{\partial \bparticle_j}\right)^2 \quad \mbox{from definition above and the Frobenius norm}\\
        & = \sum_{i=1}^d\sum_{j=1}^d \left(\langle\hat{\phi}_i(\cdot),\partial \kappa(\bparticle,\cdot)/\partial \bparticle_j \rangle_{\mathcal{H}_0}\right)^2 \quad\mbox{by Theorem 1 of \cite{zhou_derivative_2008}}  %
    \\
    & \leq \sum_{i=1}^d \sum_{j=1}^d   \norm{\hat{\phi}_i}^2_{\mathcal{H}_0}\norm{\partial\kappa(\bparticle, \cdot)/\partial \bparticle_j}^2_{\mathcal{H}_0} \quad \mbox{by Cauchy-Schwarz}\\
    & = \norm{\hat{\phi}}^2_{\mathcal{H}^d} \nabla_{\bparticle, \bparticle'}\kappa(\bparticle, \bparticle') \\
    & = \mathbb{D}(q_t,p)^2 \nabla_{\bparticle, \bparticle'}\kappa(\bparticle, \bparticle').
\end{align*}

Finally, putting terms C1 and C2 together, \eqref{equn:ProofProblem} now becomes
\begin{align*}
    \KL(q_{t+1} || p) - \KL (q_{t}|| p) & \leq -\epsilon \mathbb{D}(q_t, p)^2 +\epsilon^2\mathbb{D}(q_t,p)^2 \mathbb{E}_{\bparticle\sim q_t}\left[\nabla_{\bparticle, \bparticle'}\kappa(\bparticle, \bparticle)\right]+ \frac{\epsilon^2L}{2}\mathbb{D}(q_t , p)^2  \mathbb{E}_{\bparticle\sim q_t}\left[\kappa(\bparticle, \bparticle) \right] \\
    &\qquad  \\
    & = -\epsilon \mathbb{D}(q_t , p)^2 \left(1-\epsilon\mathbb{E}_{\bparticle\sim q_t}\left[L\kappa(\bparticle, \bparticle)/2 + \nabla_{\bparticle, \bparticle'}\kappa(\bparticle, \bparticle) \right]\right).
\end{align*}

\section{Kernel expressions}\label{app:kernelExpressions}

\begin{table}[!ht]
\centering
\caption{Explicit forms of the kernels used in \secref{sec:exps:AQ} operating on an arbitrary $\x \in \mathbb{R}^d$. For Mat\'ern kernels, the respective order is given by $\nicefrac{c}{2}-2$. For notational brevity we let $\tau = \norm{\x, \x'}_2^2$.}
\label{tab:kernels}
\begin{tabular}{l|c|c}
Kernel & $k_{\btheta}(\x, \x')$ & $\btheta$  \\ \hline
Mat\'ern \nicefrac{1}{2} & $\sigma^2\exp(\nicefrac{-\tau}{\boldsymbol{\ell}})$ & $\{\sigma \in \mathbb{R}, \boldsymbol{\ell}\in \mathbb{R}^d\}$   \\
Mat\'ern \nicefrac{5}{2} & $\sigma^2(1+\nicefrac{\sqrt{5}\tau}{\boldsymbol{\ell}}+\nicefrac{\nicefrac{5}{3}\tau^2}{\boldsymbol{\ell}^2})\exp(-\nicefrac{\sqrt{5}\tau}{\boldsymbol{\ell}})$ & $\{\sigma \in \mathbb{R}, \boldsymbol{\ell}\in \mathbb{R}^d\}$   \\
Squared exponential & $\sigma^2\exp(\nicefrac{-\tau}{2\boldsymbol{\ell}^2})$ & $\{\sigma \in \mathbb{R}, \boldsymbol{\ell}\in \mathbb{R}^d\}$   \\
Polynomial ($d$-order)  & $(\sigma^2\x^{\top}\x' + \gamma)^{d}$ & $\{\sigma \in \mathbb{R}, \gamma \in \mathbb{R}\}$ \\ 
White &     $\begin{cases}
      \sigma & \text{if $\x = \x'$}\\
      0 & \text{otherwise}
    \end{cases}$   & $\{\sigma \in \mathbb{R}\}$
    \end{tabular}
\end{table}

\clearpage

\section{Further experimental details}

\subsection{UCI datasets}\label{app:UCIDatasets}

\begin{table}[ht]
\centering
\caption{Full list of the UCI datasets used in \secref{sec:exps:UCI}. $N$ corresponds the number of observations, whilst the dimension column quantifies the dimensionality of the inputs. For datasets with a binary target, we also report the dataset's imbalance through the proportion of observations whereby the target is positive i.e. 1.}
\label{tab:regDatasets}
\begin{tabular}{lccc}
\toprule
               Dataset &        N &  Dimension & Positive-proportion\\
\midrule
    airfoil & 1503 & 5 & - \\
    autompg & 392 & 7 &  -\\
    blood &   748 &         5 &               23.8\% \\
    boston & 506 & 13 &  - \\
    breast-cancer &   286 &        10 &              29.72\% \\
    challenger & 23 & 4 &  - \\
    concrete & 1030 & 8 &  - \\
    concreteslump & 103 & 7 &  - \\
    fertility &   100 &        10 &               12.0\% \\
    gas & 2565 & 128 &  - \\
    hepatitis &   155 &        20 &              79.35\% \\
    machine & 209 & 7 &  - \\
    mammographic &   961 &         6 &              46.31\% \\
    parkinsons & 5875 & 20 &  - \\           
    servo & 167 & 4 &  - \\
    skillcraft & 3338 & 19 &  -  \\
    spectf &   267 &        45 &               79.4\% \\
    winered & 1599 & 11 &  - \\
    winewhite & 4898 & 11 &  - \\
\bottomrule
\end{tabular}
\end{table}

\subsection{Full UCI results}\label{app:fullUCILL}

\begin{table}[H]
\centering
\begingroup\scriptsize
\caption{Full set of test log-likelihood values for the datasets used in \secref{sec:exps:UCI}.\label{tab:fullUCITestLL}} 
\begin{tabular}{lcccccc}
  \hline
Dataset & SteinGP2 & SteinGP5 & SteinGP10 & SteinGP20 & VI & ML/HMC \\ 
  \hline
  Airfoil & $\mathbf{0.06\pm 0.04}$ & $ 0.06  \pm 0.04 $ & $ 0.05  \pm 0.06 $ & $ 0.05  \pm 0.05 $ & $ 0.03  \pm 0.03 $ & $ 0.03  \pm 0.03 $ \\ 
  Autompg & $ -0.39  \pm 0.09 $ & $ -0.39  \pm 0.09 $ & $ -0.39  \pm 0.09 $ & $ -0.4  \pm 0.09 $ & $\mathbf{-0.39\pm 0.07}$ & $ -0.39  \pm 0.07 $ \\ 
  Blood & $ -0.6  \pm 0.05 $ & $ -0.6  \pm 0.04 $ & $ -0.6  \pm 0.05 $ & $ -0.61  \pm 0.04 $ & $\mathbf{-0.51\pm 0.05}$ & $ -0.52  \pm 0.06 $ \\ 
  Boston & $ -0.3  \pm 0.12 $ & $\mathbf{-0.28\pm 0.11}$ & $ -0.3  \pm 0.12 $ & $ -0.3  \pm 0.13 $ & $ -0.31  \pm 0.13 $ & $ -0.31  \pm 0.13 $ \\ 
  Breast Cancer &  $-0.08 \pm 0.04$ &  $-0.08 \pm 0.02$ & $-0.08 \pm 0.02$ & $\mathbf{-0.08 \pm 0.01}$ & $-0.65 \pm 0.09$ & $\mathbf{-0.08 \pm 0.04}$ \\
  Challenger & $ -1.53  \pm 0.45 $  & $ -1.52  \pm 0.43 $ & $\mathbf{-1.46  \pm 0.32}$  & $-1.53  \pm 0.41$   & $ -1.51  \pm 0.3 $ & $-1.51\pm 0.3$ \\ 
  Concrete & $ -0.25  \pm 0.07 $ & $ -0.25  \pm 0.07 $ & $ -0.25  \pm 0.07 $ & $ -0.25  \pm 0.07 $ & $\mathbf{-0.24\pm 0.05}$ & $ -0.24  \pm 0.05 $ \\ 
  Concreteslump & $1.08\pm 0.39$ & $ 1.07  \pm 0.41 $ & $ \mathbf{1.06  \pm 0.4} $ & $ 1.08  \pm 0.39 $ & $ 0.13  \pm 1.14 $ & $ 0.13  \pm 1.14 $ \\ 
  Fertility &  $-0.44 \pm 0.03$ &  $-0.44 \pm 0.03$ & $-0.43 \pm 0.02$ & $\mathbf{-0.42 \pm 0.02}$ & $-0.70 \pm 0.08$ & $-0.54 \pm 0.02$ \\
  Gas & $ 0.88  \pm 0.11 $ & $ 0.88  \pm 0.11 $ & $\mathbf{0.89\pm 0.1}$ & $ 0.88  \pm 0.11 $ & $ 0.79  \pm 0.11 $ & $ 0.79  \pm 0.11 $ \\ 
  Hepatitis & $ -0.41  \pm 0.07 $ & $ -0.41  \pm 0.07 $ & $ -0.42  \pm 0.07 $ & $\mathbf{-0.4\pm 0.07}$ & $ -0.69  \pm 0 $ & $ -0.44  \pm 0.04 $ \\ 
  Machine & $\mathbf{-0.51\pm 0.09}$ & $ -0.52  \pm 0.08 $ & $ -0.52  \pm 0.08 $ & $ -0.52  \pm 0.08 $ & $ -0.52  \pm 0.07 $ & $ -0.52  \pm 0.07 $ \\ 
  Mammographic & $ -0.37  \pm 0.03 $ & $ -0.37  \pm 0.03 $ & $ -0.37  \pm 0.03 $ & $ -0.37  \pm 0.03 $ & $-0.39\pm 0.03$ & $ -0.38  \pm 0.04 $ \\ 
  Parkinsons & $ 4.12  \pm 0.05 $ & $ 4.12  \pm 0.05 $ & $\mathbf{4.14\pm 0.03}$ & $ 4.13  \pm 0.06 $ & $ 3.95  \pm 0.04 $ & $ 3.95  \pm 0.04 $ \\ 
  Servo & $ -0.48  \pm 0.04 $ & $ -0.43  \pm 0.05 $ & $ -0.41  \pm 0.11 $ & $ -0.41  \pm 0.21 $ & $ -0.39  \pm 0.1 $ & $\mathbf{-0.39\pm 0.1}$ \\  
  Skillcraft & $\mathbf{-0.99\pm 0.02}$ & $ -0.99  \pm 0.02 $ & $ -0.99  \pm 0.02 $ & $ -0.99  \pm 0.02 $ & $ -1.01  \pm 0.02 $ & $ -1.01  \pm 0.02 $ \\ 
  Spectf & $-0.26\pm 0.01$ & $ -0.26  \pm 0.01 $ & $ -0.26  \pm 0.01 $ & $ \mathbf{-0.26  \pm 0.01} $ & $ -0.69  \pm 0 $ & $ -0.68  \pm 0.03 $ \\ 
  Winered & $ -1.17  \pm 0.03 $ & $ -1.17  \pm 0.03 $ & $ -1.17  \pm 0.03 $ & $ -1.17  \pm 0.03 $ & $\mathbf{-1.16\pm 0.03}$ & $ -1.16  \pm 0.03 $ \\ 
  Winewhite & $ 0.56  \pm 0.05 $ & $ 0.57  \pm 0.05 $ & $\mathbf{0.57\pm 0.05}$ & $ 0.57  \pm 0.05 $ & $ 0.49  \pm 0.05 $ & $ 0.55  \pm 0.05 $ \\ 
\hline
\end{tabular}
\endgroup
\end{table}

\subsection{Computational runtimes}\label{app:UCIRuntimes}

\begin{table}[ht]
\centering
\scriptsize
\caption{Computational runtimes reported in seconds for each model assessed in \secref{sec:exps:UCI}.} 
\label{tab:reg_time}

\resizebox{\textwidth}{!}{%
\begin{tabular}{lccccccc}
  \hline
Dataset & SteinGP2 & SteinGP5 & SteinGP10 & SteinGP20 & VI & ML & HMC \\ 
  \hline
  Airfoil & $ 39.96  \pm 1.25 $ & $ 74.33  \pm 2.51 $ & $ 129.81  \pm 5.36 $ & $ 240.74  \pm 2.93 $ & $ 61.48  \pm 5.11 $ & $\mathbf{21\pm 1.87}$  & - \\ 
  Autompg & $ \mathbf{10.09  \pm 1.49} $ & $ 19.91  \pm 3.53 $ & $ 36.11  \pm 10.17 $ & $ 63.93  \pm 8.31 $ & $ 29.02  \pm 0.46 $ & $11.97\pm 0.21$ & -  \\ 
  Blood & $ 57.71  \pm 0.26 $ & $ 111.41  \pm 2.06 $ & $ 202.47  \pm 2.35 $ & $ 381.57  \pm 2.89 $ & $ 33.19  \pm 1.42 $ & -  & $ 171.49  \pm 0.67 $ \\ 
  Boston & $ 24.5  \pm 1.7 $ & $ 49.22  \pm 7.06 $ & $ 75.34  \pm 7.17 $ & $ 151.41  \pm 22.14 $ & $ 40.29  \pm 1.81 $ & $\mathbf{14.05\pm 0.64}$ & -  \\ 
  Breast cancer & $ 12  \pm 0.49 $ & $ 24.01  \pm 1.27 $ & $ 38.3  \pm 0.68 $ & $ 70.21  \pm 0.9 $ & $ 8.82  \pm 0.07 $ & -  & $ 141.35  \pm 0.71 $ \\
  Challenger & $ 9.58  \pm 0.31 $ & $ 16.40  \pm 0.5 $ & $ 28.42  \pm 0.38 $ & $ 52.7  \pm 0.63 $ & $ 8.4  \pm 0.03 $ & $\mathbf{2.55\pm 0.07}$ & -  \\ 
  Concrete & $\mathbf{9.67\pm 1.18}$ & $ 65.21  \pm 0.97 $ & $ 85.99  \pm 21.05 $ & $ 93.73  \pm 29.91 $ & $ 39.82  \pm 0.81 $ & $ 13.89  \pm 0.33 $ & -  \\ 
  Concreteslump & $ 32.37  \pm 1.91 $ & $ 56.66  \pm 5.66 $ & $ 95.8  \pm 10.16 $ & $ 175.98  \pm 14.09 $ & $ 35.92  \pm 22.12 $ & $\mathbf{12.51\pm 8.01}$ & -  \\ 
  Fertility & $ 9.91  \pm 0.37 $ & $ 18.9  \pm 0.45 $ & $ 33.2  \pm 0.62 $ & $ 60.82  \pm 1.8 $ & $ 8.71  \pm 0.09 $ & -  &  $ 141.6  \pm 0.42 $ \\ 
  Gas & $\mathbf{52.82\pm 0.31}$ & $ 107.62  \pm 1.29 $ & $ 199.79  \pm 2.27 $ & $ 384.25  \pm 3.74 $ & $ 245.96  \pm 8.47 $ & $ 102.32  \pm 3.21 $ & -  \\ 
  Hepatitis & $ 13.4  \pm 0.47 $ & $ 25.84  \pm 0.99 $ & $ 44.56  \pm 1.06 $ & $ 81.65  \pm 2.06 $ & $ 8.78  \pm 0.05 $ & -  & $ 141.44  \pm 0.63 $ \\ 
  Machine & $ 18.23  \pm 1.5 $ & $ 29.56  \pm 2.34 $ & $ 50.07  \pm 3.13 $ & $ 97.68  \pm 7.14 $ & $ 27.98  \pm 0.93 $ & $\mathbf{9.7\pm 0.34}$ & -  \\ 
  Mammographic & $ 57.69  \pm 0.59 $ & $ 111.91  \pm 1.67 $ & $ 206.66  \pm 4.12 $ & $ 391.31  \pm 5.97 $ & $ 42.78  \pm 2.19 $ & -  & $ 190.2  \pm 0.75 $ \\ 
  Parkinsons & $\mathbf{201.64\pm 1.14}$ & $ 468.34  \pm 1.49 $ & $ 908.08  \pm 2.13 $ & $ 1783.14  \pm 4.22 $ & $ 2926.05  \pm 3.04 $ & $ 675.27  \pm 1.47 $ & -  \\ 
  Servo & $\mathbf{6.98\pm 0.73}$ & $ 42.76  \pm 21.21 $ & $ 38.15  \pm 23.22 $ & $ 61.65  \pm 15.05 $ & $ 23.94  \pm 1.52 $ & $ 8.21  \pm 0.56 $ & -  \\ 
  Skillcraft & $ 68.63  \pm 1.24 $ & $ 145.9  \pm 0.67 $ & $ 273.3  \pm 0.57 $ & $ 530.47  \pm 1.92 $ & $ 190.17  \pm 5.35 $ & $\mathbf{53.96\pm 1.38}$ & -  \\ 
  Spectf & $ 16.09  \pm 0.56 $ & $ 30.87  \pm 0.97 $ & $ 53.99  \pm 1.49 $ & $ 102.61  \pm 3.61 $ & $ 9.16  \pm 0.23 $ & -  & $ 142.41  \pm 0.52 $ \\ 
  Wine & $\mathbf{15.73\pm 0.77}$ & $ 61.99  \pm 15.93 $ & $ 69.02  \pm 7.51 $ & $ 100.62  \pm 9.2 $ & $ 48.45  \pm 4.25 $ & $ 17.46  \pm 1.31 $ & -  \\ 
  Winered & $ 16.93  \pm 3.07 $ & $ 46.44  \pm 8.93 $ & $ 55.55  \pm 8.86 $ & $ 95.14  \pm 16.87 $ & $ 28.56  \pm 2.52 $ & $\mathbf{9.83\pm 0.91}$ & -  \\ 
  Winewhite & $\mathbf{142.04\pm 0.59}$ & $ 320.33  \pm 1.46 $ & $ 616.94  \pm 1.08 $ & $ 1210.34  \pm 1.65 $ & $ 1232.24  \pm 7.84 $ & $ 315.16  \pm 1.63 $ & -  \\
   \hline
\end{tabular}
}
\end{table}

\FloatBarrier

\clearpage

\subsection{Multimodal data}\label{app:Multimodal}

\begin{figure}[ht]
    \centering
    \includegraphics[width=0.9\textwidth, height=0.3\textheight]{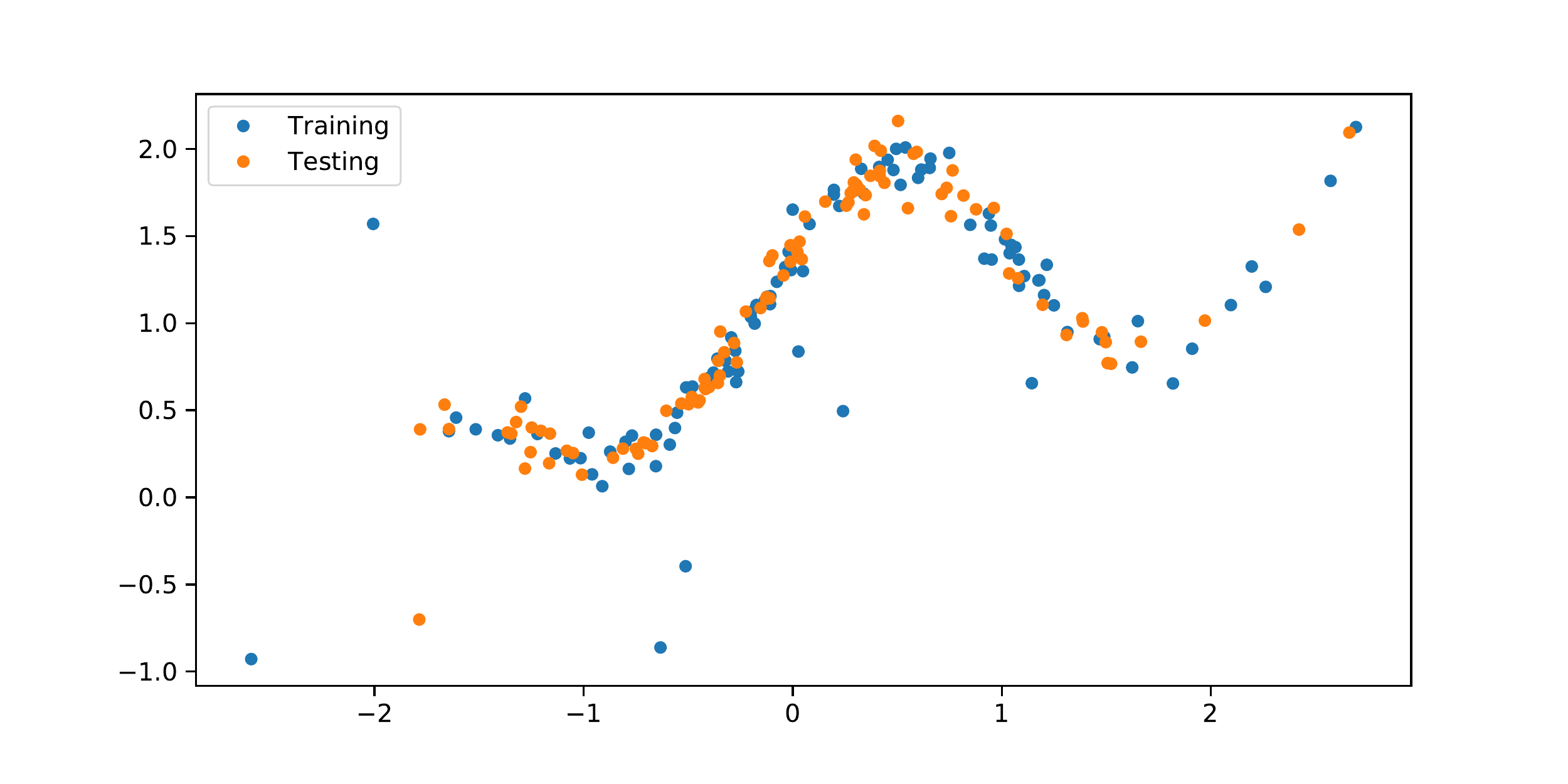}
    \caption{The multimodal dataset from \cite{neal_monte_1997} that is used in \secref{sec:exps:multimodal}.}
    \label{fig:multimodalDataset}
\end{figure}

\FloatBarrier

\subsection{Air quality stations}

\begin{figure}[H]
    \centering
    \includegraphics[width=0.55\textwidth, height=0.5\textheight]{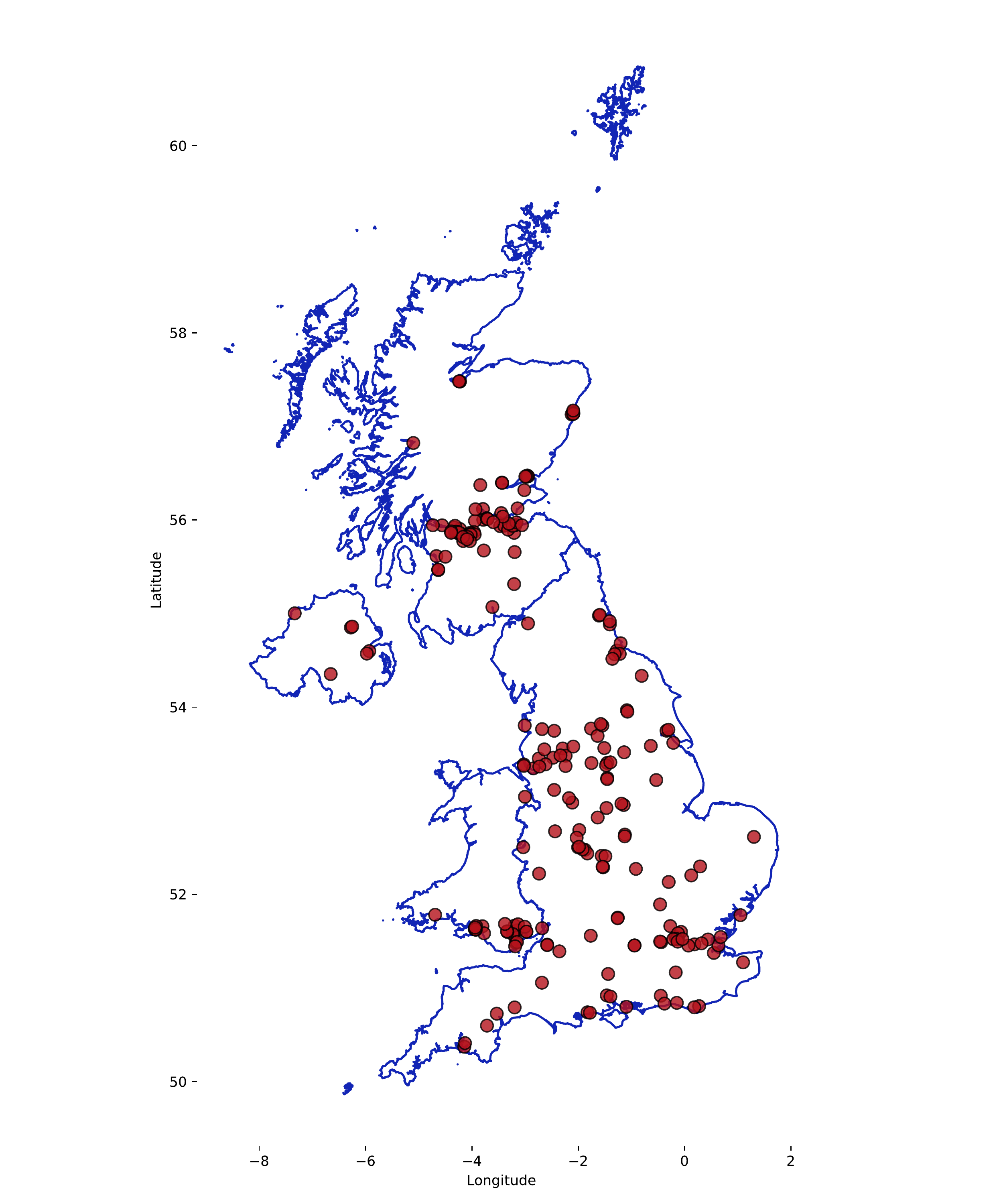}
    \caption{Spatial locations of the AURN \notwo measurement stations described in \secref{sec:exps:AQ}.}
    \label{fig:stationLocation}
\end{figure}

\FloatBarrier

\subsection{Temporal behaviour of air quality}\label{app:AQ:Temporal}

\begin{figure}[H]
    \centering
    \includegraphics[width=0.7\textwidth, height=0.25\textheight]{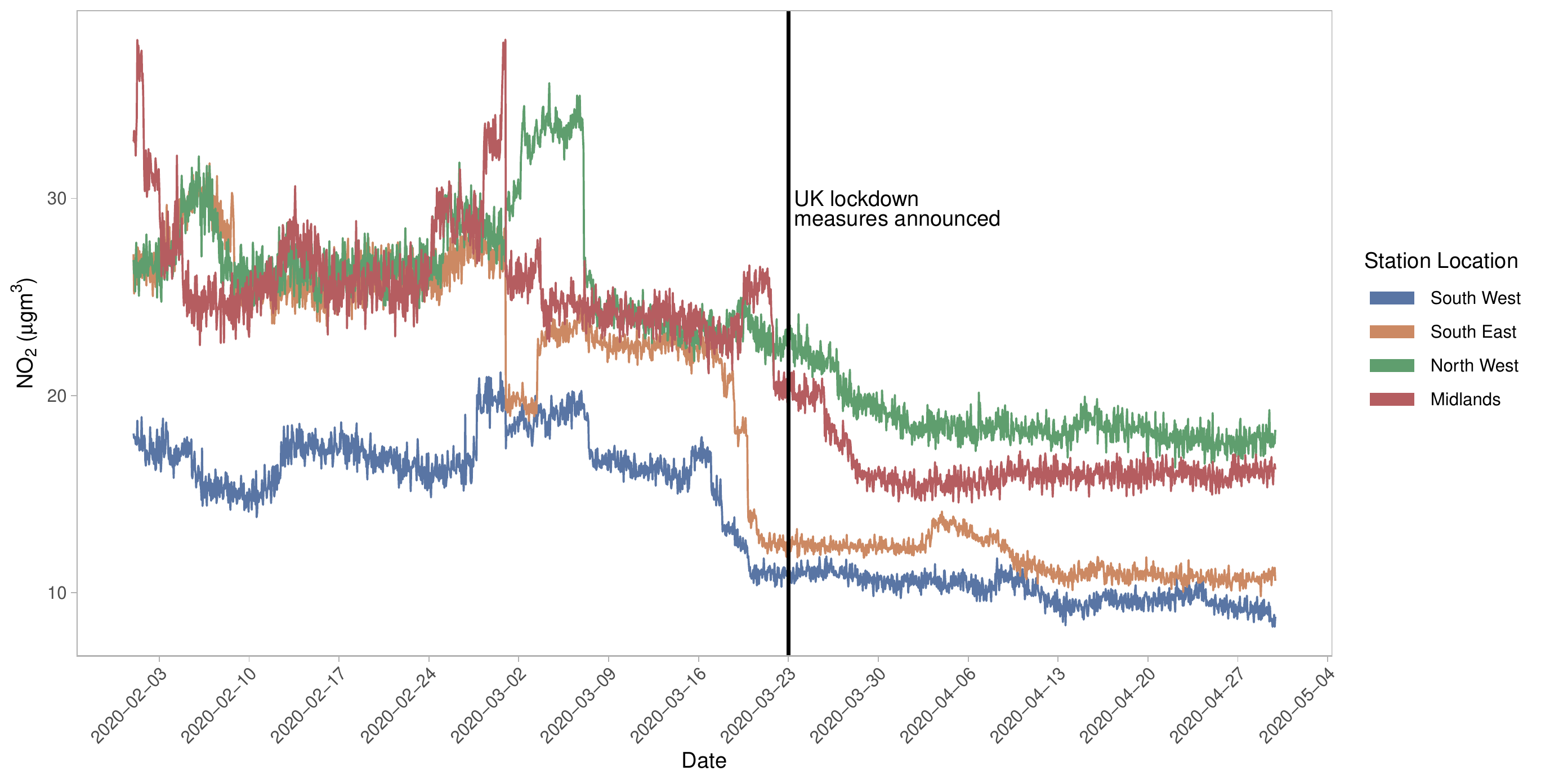}
    \caption{Example of the complex nonstationary behaviour of NO$_2$ levels over time. Four time series originating from four stations across the UK are visualised with hourly values from February 1$^{\text{st}}$ through to April 30$^{\text{th}}$.}
    \label{fig:my_label}
\end{figure}

\FloatBarrier

\subsection{Spatial interpolation predictive results}

\begin{table}[h]
\centering
\caption{Spatial interpolation of a SteinGP with 30 particles and a stochastic \gls{VI} \gls{GP}. Here, stations within (-2.2$^{o}$, 52$^{o}$), (-1$^{o}$, 54$^{o}$) are used for testing.}
\label{tab:aqSpatialResults}
\begin{tabular}{lcc}
\hline
               & SteinGP   & SVI       \\ \hline
RMSE           & $0.97 $ & $1.00$ \\
Log-likelihood & $-1.467 $ &  $-1.462$        \\
\hline

\end{tabular}
\end{table}

\FloatBarrier

\section{Demo implementation}\label{sec:app:demo}

\subsection{Synthetic example}

The accompanying code to this paper has been designed to integrate with GPFlow 2, and can be run through the following commands.

\begin{python}
from steingp import SteinGPR, RBF, Median, SVGD
import numpy as np 
import gpflow

# Build data
X = np.random.uniform(-5, 5, 100).reshape(-1,1)
y = np.sin(x)

# Define model
kernel = gpflow.kernels.SquaredExponential()
model = SteinGPR((X, y), kernel)

# Fit 
opt = SVGD(model, RBF(bandwidth=Median()), n_particles=5)
opt.run(iterations = 1000)

# Predict
Xtest = np.linspace(-5, 5, 500).reshape(-1, 1)
theta = opt.get_particles()
posterior_samples = model.predict(Xtest, theta, n_samples=5)
\end{python}

\clearpage

\subsection{Air quality example}\label{app:demo:AQ}

\begin{python}
from steingp import StochSteinGPR, RBF, Median, SVGD
import numpy as np 
import pandas as pd
import tensorflow as tf
from gpflow.kernels import Matern52, Polynomial, White
from gpflow.likelihoods import Gaussian
from scipy.cluster.vq import kmeans 2

# Load data
AIR_QUALITY_DATA_PATH = ""
aq = pd.read_csv(AIR_QUALITY_DATA_PATH)
X, y = aq[:, :3].values, aq[:, 3].values

# Spatial kernel
k1 = Matern52(lengthscales=[np.sqrt(2)]*2, active_dims = [0, 1])
# Temporal kernel
k2 = Polynomial(active_dims=[2]) * Matern52(active_dims=[2])
# Gather terms
kern = k1 * k2 * White()

# Inducing points
Z = kmeans2(X, k=500, minit='points')[0]

# Define model
model = StochSteinGPR(kernel, likelihood = Gaussian(), inudcing_variables=Z)

# Setup data
minibatch_size = 128
N = X.shape[0]
train_dataset = tf.data.Dataset.from_tensor_slices((X, y)).repeat().shuffle(N)
train_iter = iter(train_dataset.batch(minibatch_size))

# Fit 
opt = SVGD(model, train_iter, RBF(bandwidth=Median()), n_particles=30)
opt.run(iterations = 1000)

# Predict
theta = opt.get_particles()
posterior_samples = model.predict(X, theta, n_samples=5)
\end{python}

\section{Training details}\label{sec:app:expDetails}
\paragraph{GP kernels} For the regression and classification experiments in \secref{sec:exps:UCI}, an ARD RBF kernel was used. The same RBF kernel was used for the multimodal example in \secref{sec:exps:multimodal}. A Matern32 kernel was used for the spatial inputs in in \secref{sec:exps:AQ}, whilst a third-order polynomial kernel was used for the temporal inputs. In all experiments, the kernel's lengthscale for each dimension was initialised at the square root of the data's dimensionality. Further, the kernel's variance was initialised to equal 1.0.

\paragraph{SVGD kernels} In all experiments we use an RBF kernel to compute \eqref{equn:SVGD_update}. The kernel's lengthscale is estimated at each iteration of the optimisation procedure using the median rule, as per \cite{liu_stein_2016}.

\paragraph{Likelihoods} For the Gaussian likelihood functions used in \secref{sec:exps:UCI} and \ref{sec:exps:multimodal}, the variance parameter was initialised to 1.0. The same Gaussian likelihood is used in \ref{sec:exps:AQ} but no initialisation is required due to the use of priors.

\paragraph{Particle initialisation} The number of particles used in the \gls{svgd} scheme is explicitly reported in the respective results. Particle initialisation is carried out by making a random draw from the respective parameter's prior distributions where applicable, otherwise a draw is made from the uniform distribution on [0,1]. 

\paragraph{Prior distributions} For GP models fitted using either \gls{HMC} or \gls{svgd} we place the same priors on all parameters. Unless explicitly specified, a Gamma distribution parameterised with a unit shape parameter and a scale parameter of 2 is used as the prior for all lengthscale and variance parameters.

\paragraph{Parameter constraints} For all parameters where positivity is a constraint (i.e. variance), the softplus transformation is applied with a clipping of $10^{-6}$, as is the default in GPFlow. Optimisation is then conducted on the constrained parameter, however, we report the re-transformed parameter i.e. the unconstrained representation.

\paragraph{Optimisation} For all variational models, natural gradients are used to optimise the variational parameters with a stepsize of 0.1. For the kernel hyperparameters and likelihood parameters (where applicable) in both the variational models and models fitted using maximum likelihood, the Adam optimiser \citep{kingma_adam:_2015} was used. In \secref{sec:exps:UCI} a step-size parameter of 0.05 is used, instead of the default recommendation of 0.001, as this was found to give faster optimisation at no detriment to the model's predictive accuracy. For the multimodal and air quality experiments in section \secref{sec:exps:multimodal} and \ref{sec:exps:AQ}, the default Adam learning rate of 0.001 was used.

\paragraph{Data availability} All datasets used in \secref{sec:exps:UCI} are available at \\ {\tt https://github.com/RedactedForReview}, based upon the work in \cite{salimbeni_bayesian_2019}. The multimodal data is from \cite{neal_monte_1997}. The air quality data used in \secref{sec:exps:AQ} was gathered using the {\fontfamily{cmtt}\selectfont openaiR} package \citep{carslaw_openair_2012} using a script available at {\tt https://github.com/RedactedForReview}.

\paragraph{Implementation}All code for this work is written in TensorFlow \citep{abadi_tensorflow_2016} and extends the popular Gaussian process library GPFlow \citep{matthews_gpflow:_2017}. Code to replicate experiments can be found at {\tt https://github.com/RedactedForReview}.

\FloatBarrier

\end{document}